\documentclass[lettersize,journal]{IEEEtran}

\usepackage{times}
\usepackage{epsfig}
\usepackage{graphicx}
\usepackage{amsmath}
\usepackage{amssymb}
\usepackage{subfigure}
\usepackage{xcolor}
\usepackage{booktabs}
\usepackage{array}
\usepackage{multirow}
\usepackage{float}

\usepackage{cuted}
\usepackage{mathtools}
 \usepackage{caption}
 \usepackage{fontawesome5} 

\usepackage[pagebackref=true,breaklinks=true,letterpaper=true,colorlinks,bookmarks=false]{hyperref}

\begin{document}

\title{A Unified Simulation Framework for Visual and Behavioral Fidelity in Crowd Analysis}

\author{Niccol\'o Bisagno, Nicola Garau, Antonio Luigi Stefani, and Nicola Conci\\
University of Trento\\
CNIT - Consorzio Interuniversitario per le Telecomunicazioni\\
Via Sommarive, 9, 38123 Povo, Trento TN (Italy)\\
{\tt\small \{niccolo.bisagno, nicola.garau, antonioluigi.stefani, nicola.conci\}@unitn.it}}

\maketitle

\begin{abstract}
Simulation is a powerful tool to easily generate annotated data, and a highly desirable feature, especially in those domains where learning models need large training datasets. Machine learning and deep learning solutions, have proven to be extremely data-hungry and sometimes, the available real-world data are not sufficient to effectively model the given task. Despite the initial skepticism of a portion of the scientific community, the potential of simulation has been largely confirmed in many application areas, and the recent developments in terms of rendering and virtualization engines, have shown a good ability also in representing complex scenes. This includes environmental factors, such as weather conditions and surface reflectance, as well as human-related events, like human actions and behaviors.
We present a human crowd simulator, called UniCrowd, and its associated validation pipeline. We show how the simulator can generate annotated data, suitable for computer vision tasks, in particular for detection and segmentation, as well as the related applications, as crowd counting, human pose estimation, trajectory analysis and prediction, and anomaly detection.
\end{abstract}


\section{Introduction}
\label{sec:intro}

\begin{table*}
\centering
    \begin{tabular}{ccc}
    \includegraphics[width=0.32\textwidth]{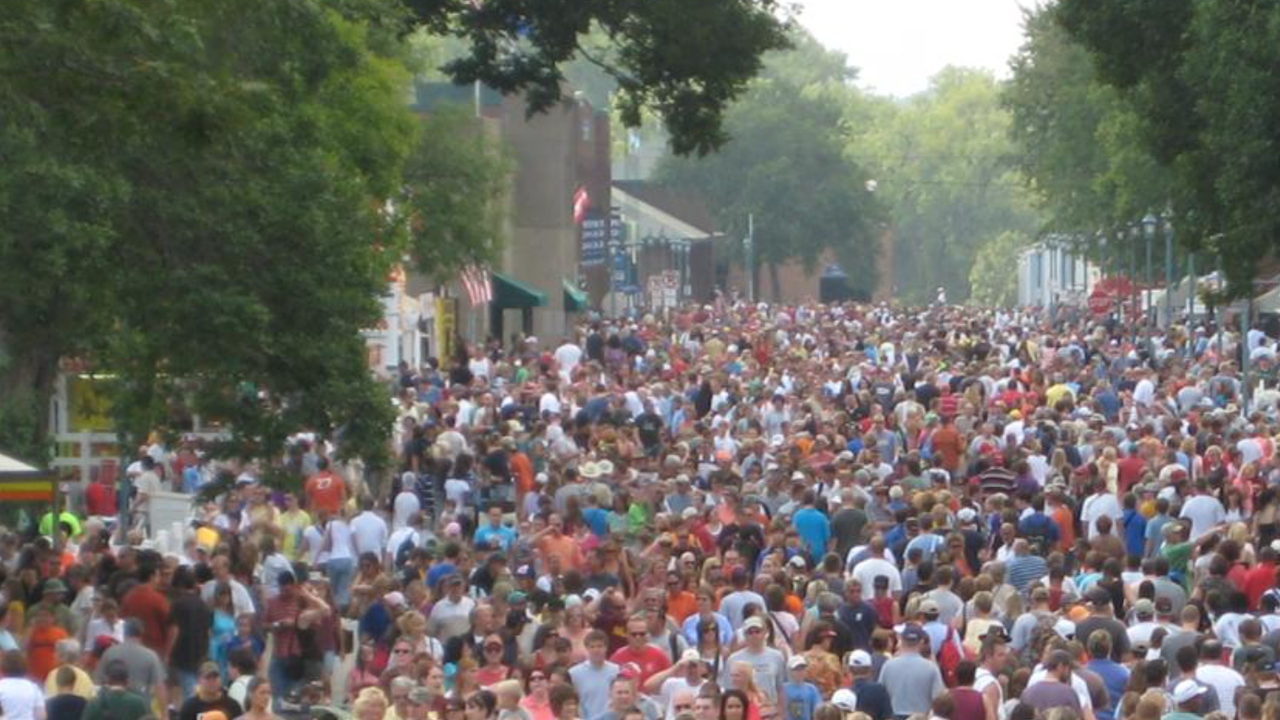} &
    \includegraphics[width=0.32\textwidth]{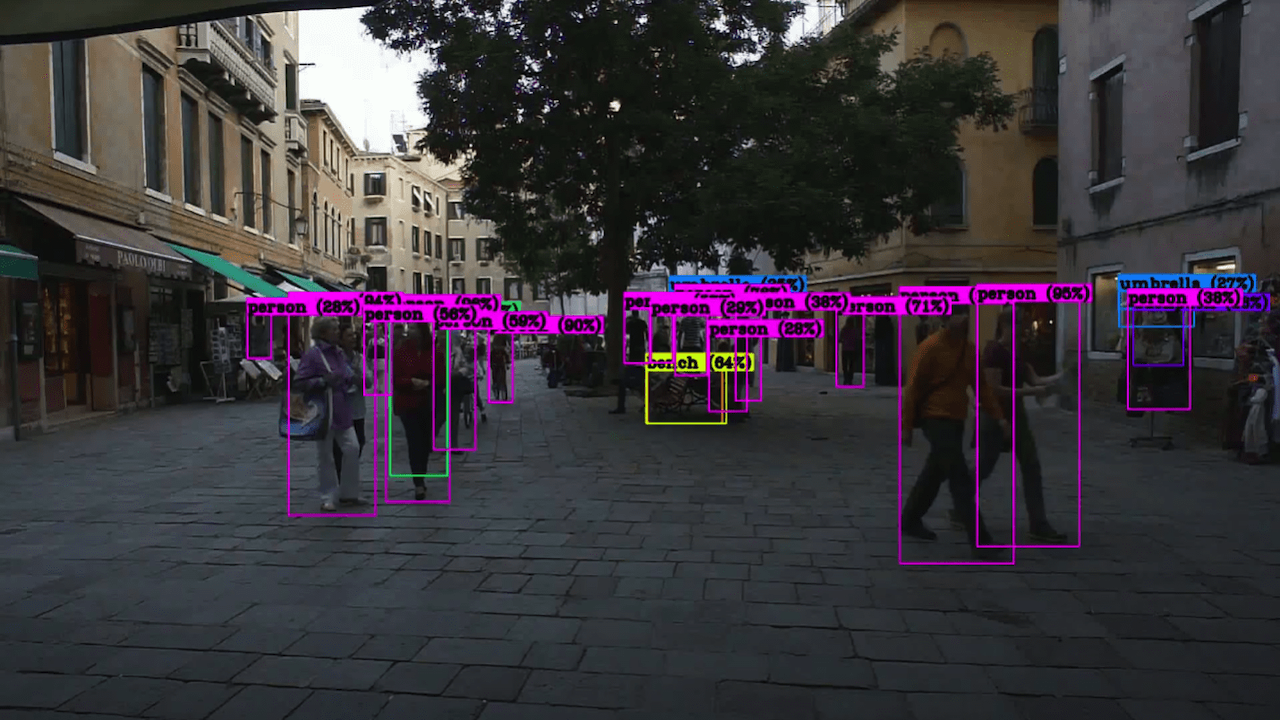} &
    \includegraphics[width=0.32\textwidth]{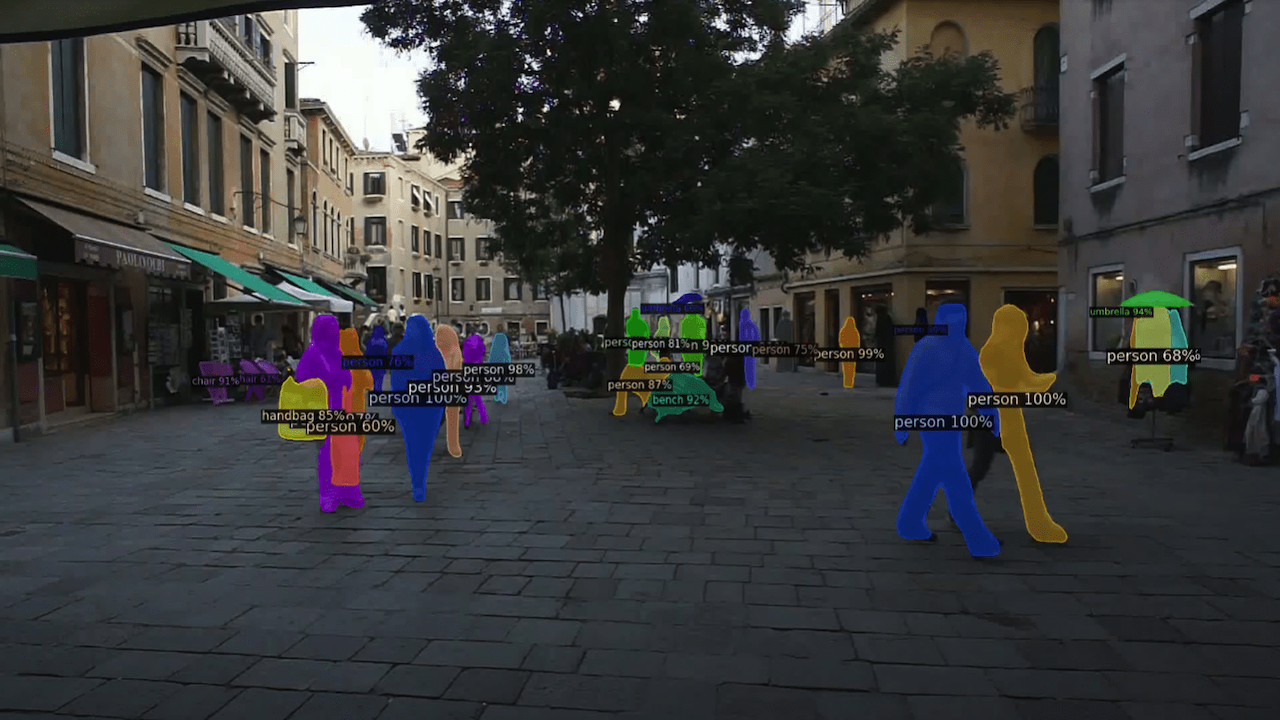} \\
     (a) & (b) & (c)\\
     \end{tabular}
    \begin{tabular}{ccc}
    \includegraphics[width=0.32\textwidth]{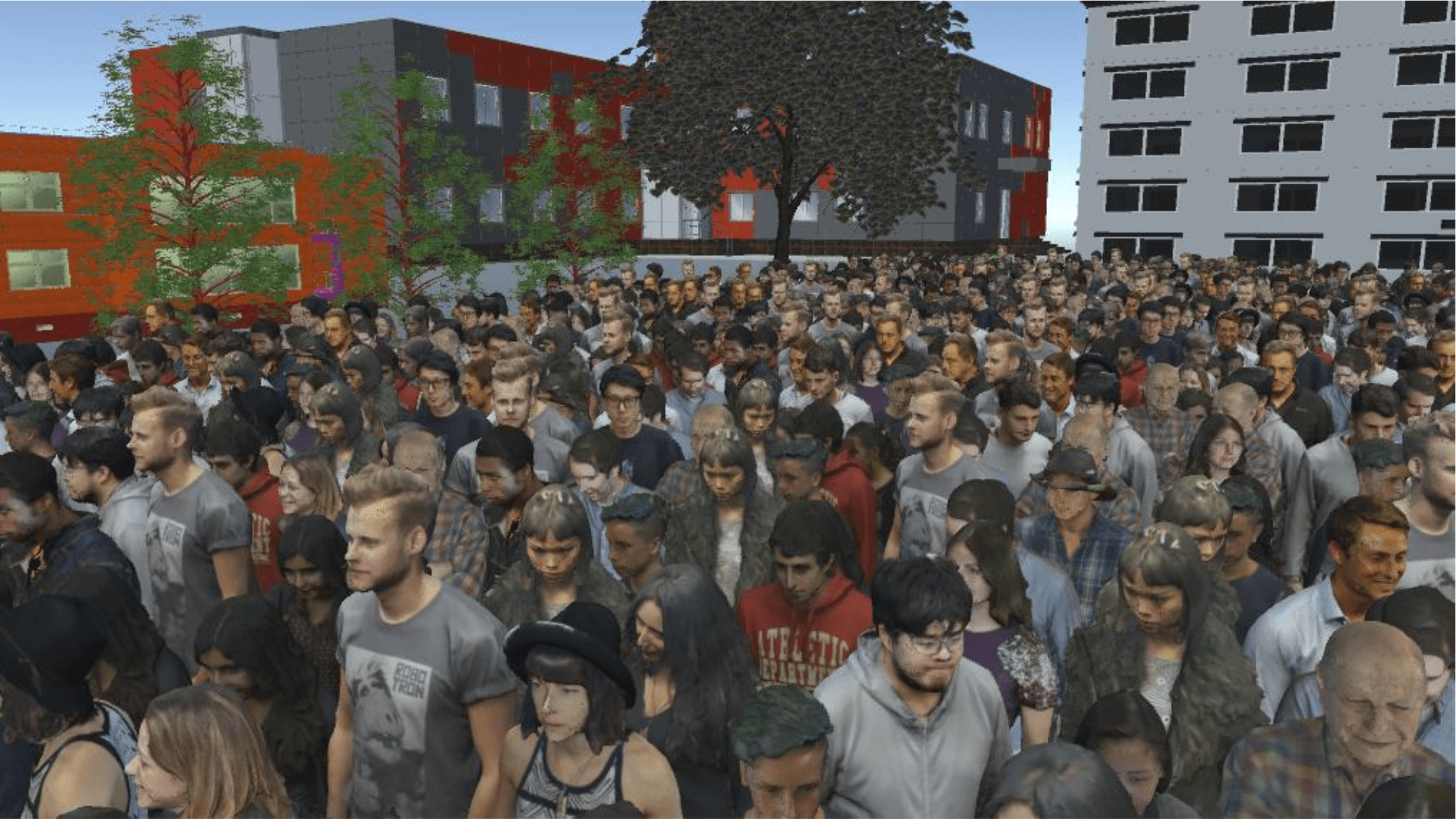} &
    \includegraphics[width=0.32\textwidth]{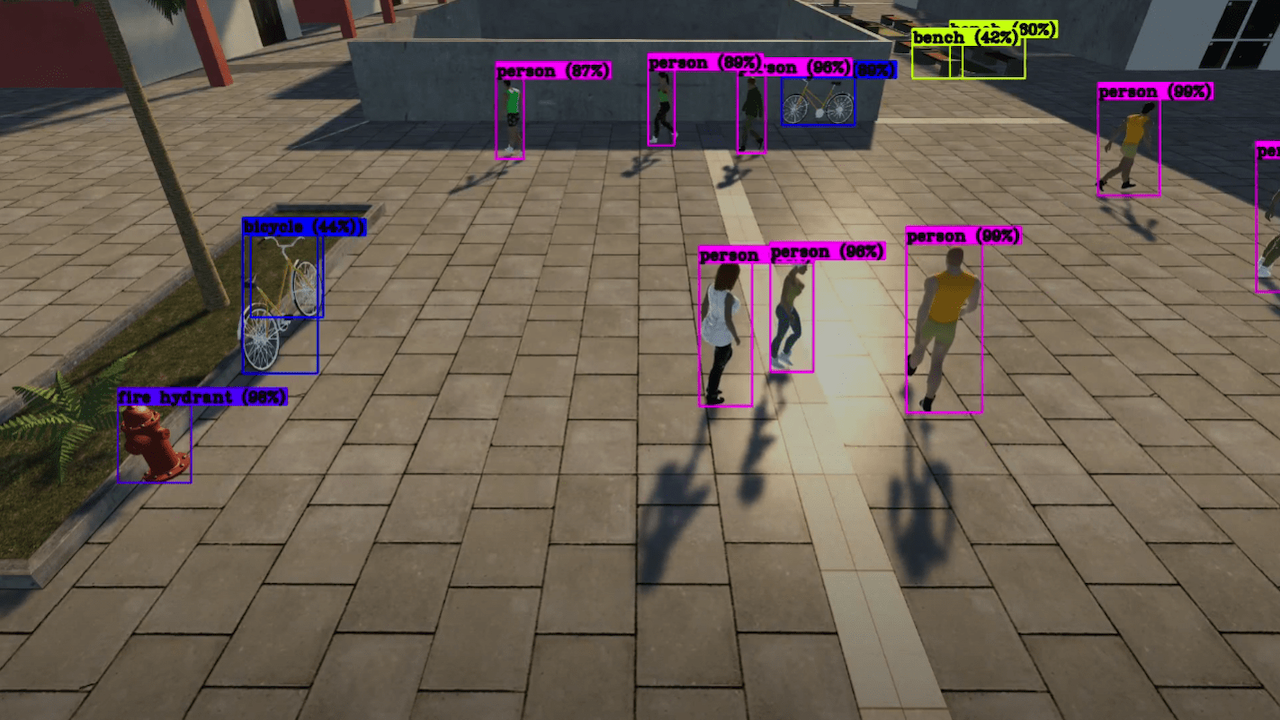} &
    \includegraphics[width=0.32\textwidth]{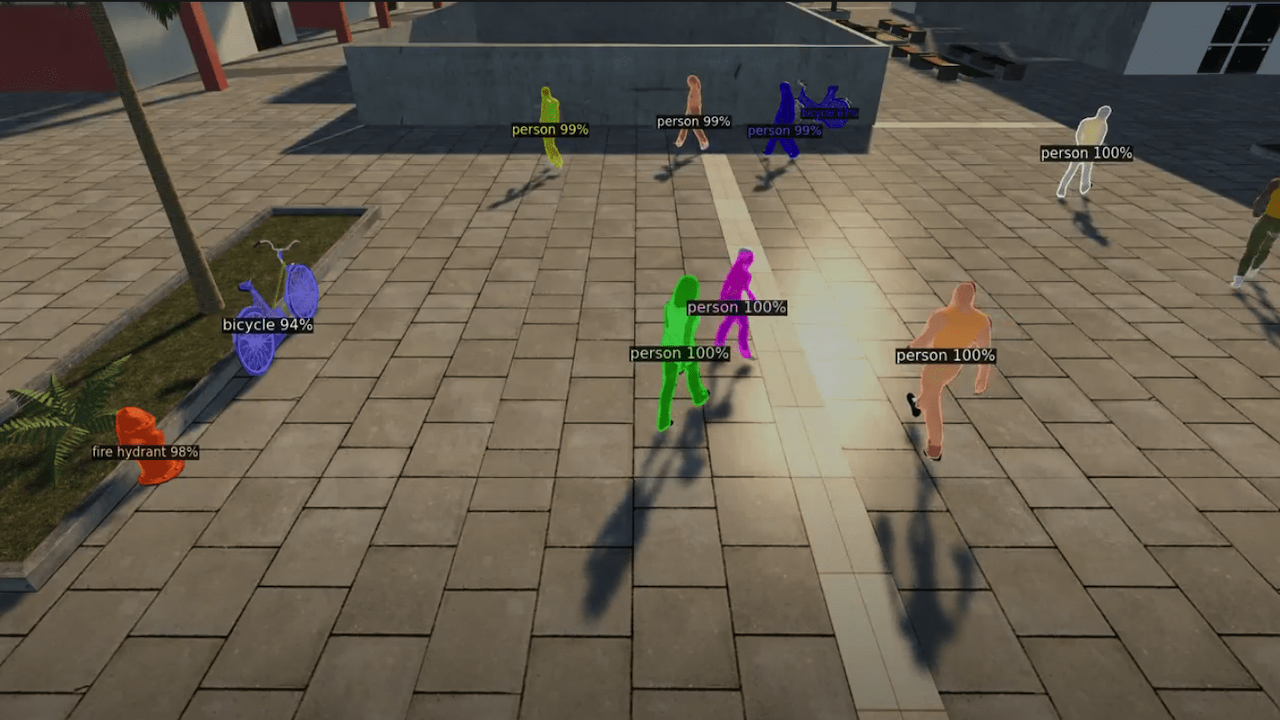} \\
     (d) & (e) & (f)\\
    \end{tabular}
    \captionof{figure}{Simulating synthetic data for crowd analysis generated with the UniCrowd simulator (second row) and real crowd data (first row). The UniCrowd simulator can generate video sequences and the ground truth data suitable for a number of tasks crowd counting (d), people detection (e) and people segmentation (f). In this work we show how having a multi-purpose simulation framework able to reproduce both behavior and appearance of crowds allows to generate effective synthetic data for multiple tasks. }
    \label{fig:application_samples}
\end{table*}

The ever increasing demand for data to train modern machine learning and deep learning algorithms appears to be unstoppable. When a given problem appears to be nearly solved, a new one takes over, showing the limitations of the previously developed architectures and the associated datasets in addressing it. This is common in well-known tasks like image classification \cite{krizhevsky2012imagenet,deng2009imagenet}, natural language processing \cite{wu2016google,bahdanau2014neural}, speech recognition  \cite{fma_dataset}, medical and ultrasound imaging \cite{zhou2021review}, remote sensing \cite{yuan2020deep}). 

To cope with this issue, researchers have come up with a very heterogeneous set of possible solutions, exploring supervised and unsupervised approaches. As for supervised learning, the basic requirement is the availability of a large collection of labelled data. However, if the amount of annotated data is scarce, supervised solutions tend to overfit, leading to poor generalisation capabilities. The literature has shown that this problem can be mitigated with a variety of regularization techniques, such as the dropout \cite{srivastava2014dropout}, batch normalization \cite{ioffe2015batch}, transfer learning between different datasets \cite{bengio2012deep}, pre-training the network on different datasets \cite{erhan2010does}, or implementing few-shot \cite{sung2018learning} and zero-shot learning \cite{xian2017zero} techniques. 
Another way to tackle the data starvation problem consists of looking at it from the data perspective: fine-tuning and data augmentation are among the most common solutions \cite{shorten2019survey}.
Through fine tuning, we teach a network that was conceived to address a certain task, to reconfigure itself to solve a different problem; with data augmentation, we acknowledge that the amount and quality of data necessary to perform the training is not sufficient, and it is necessary to apply transformations to the available samples so as to create new instances.
In both cases, this requires extensive and costly annotation sessions, generally performed manually, and validated by humans. In image processing and computer vision, the annotation cost is particularly high, since images and videos are annotated one by one, often requiring the manual selection of the relevant regions of interest. This may lead to the so-called \textit{curse of dataset annotation} \cite{xie2016semantic}: when datasets tend to be smaller, the more detailed the annotation is. Viceversa, the dataset size is prioritized over the annotation accuracy, leading to poor or inconsistent labels \cite{deng2009imagenet}. 

Some approaches also rely on unsupervised learning, which leverages the potential of not requiring labeled samples, through, for example, clustering and self-supervision \cite{henaff2021efficient}. However, unsupervised techniques tend to exhibit their inherent shortcomings in terms of accuracy, being less performing than supervised approaches.

When we refer to data, they can be collected both in controlled environments \cite{Joo_2017_TPAMI} or, as we say, \textit{in the wild}. Data acquisition in controlled environments usually delivers better annotated data, thanks to an adequate acquisition setup, though with the risk of being less representative of real-world scenarios. When working \textit{in the wild}, issues emerge, such as privacy, light conditions, overall quality of the data, need of complex setups and costly equipment.

Among the tools to address the lack of annotated data, researchers have explored the chance of replacing (or complementing) the collected samples by using simulation frameworks \cite{allain2012agoraset,fabbri2018learning,richter2016playing,shotton2013real}. 
Simulators provide accurate ground truth data, reducing the time required for annotation and enabling the collection of bigger and exhaustive datasets.
However, relying on simulators might introduce additional issues, as the generated (synthetic) data must closely resemble their real counterpart. 

Simulators should then ensure that the generated data comply with \textit{visual fidelity}, and, when humans or moving agents are involved, \textit{behavioral fidelity}.

\textit{Visual Fidelity} assesses how close the synthetic scene visually resembles a real-world one, as captured by a real camera. A given algorithm should perform comparably when trained with either real or synthetic data generated with a simulator. Thus, one of the goals of a simulator is to provide data that allow a (possibly) seamless transfer between the synthetic domain and the real one.

\textit{Behavioral Fidelity} refers to the ability of the simulator to model the dynamics of the target domain in a coherent and realistic way. This requires the observation of real-world behavioral dynamics, which must be consequently modeled in a synthetic scene. This is particularly relevant when dealing with videos, to guarantee temporal consistency across successive frames.

With this respect, crowd analysis provides a rich and diversified use case, in which simulators can play a relevant role: the scene should replicate the appearance of a crowd, which consists of multiple subjects exhibiting different appearance and behaviors. These elements imply fulfilling the requirements of both \textit{visual fidelity} and \textit{behavioral fidelity}, simulating and modeling the diversity of motion patterns, as well as the ongoing social interactions.

In this work we present a simulation framework, which is suitable for image processing and computer vision tasks, in particular for the monitoring and surveillance of crowded scenes. 

Compared to the existing available synthetic datasets \cite{fabbri2018learning,Butler:ECCV:2012,mayer2018makes,ros2016synthia,gaidon2016virtual}, we provide the possibility to generate data, in a fully customisable fashion, in terms of number of cameras, their position and resolution, as well as the environment layout and features.

\begin{figure*}
    \centering
    \subfigure[]{\includegraphics[height=0.165\textwidth]{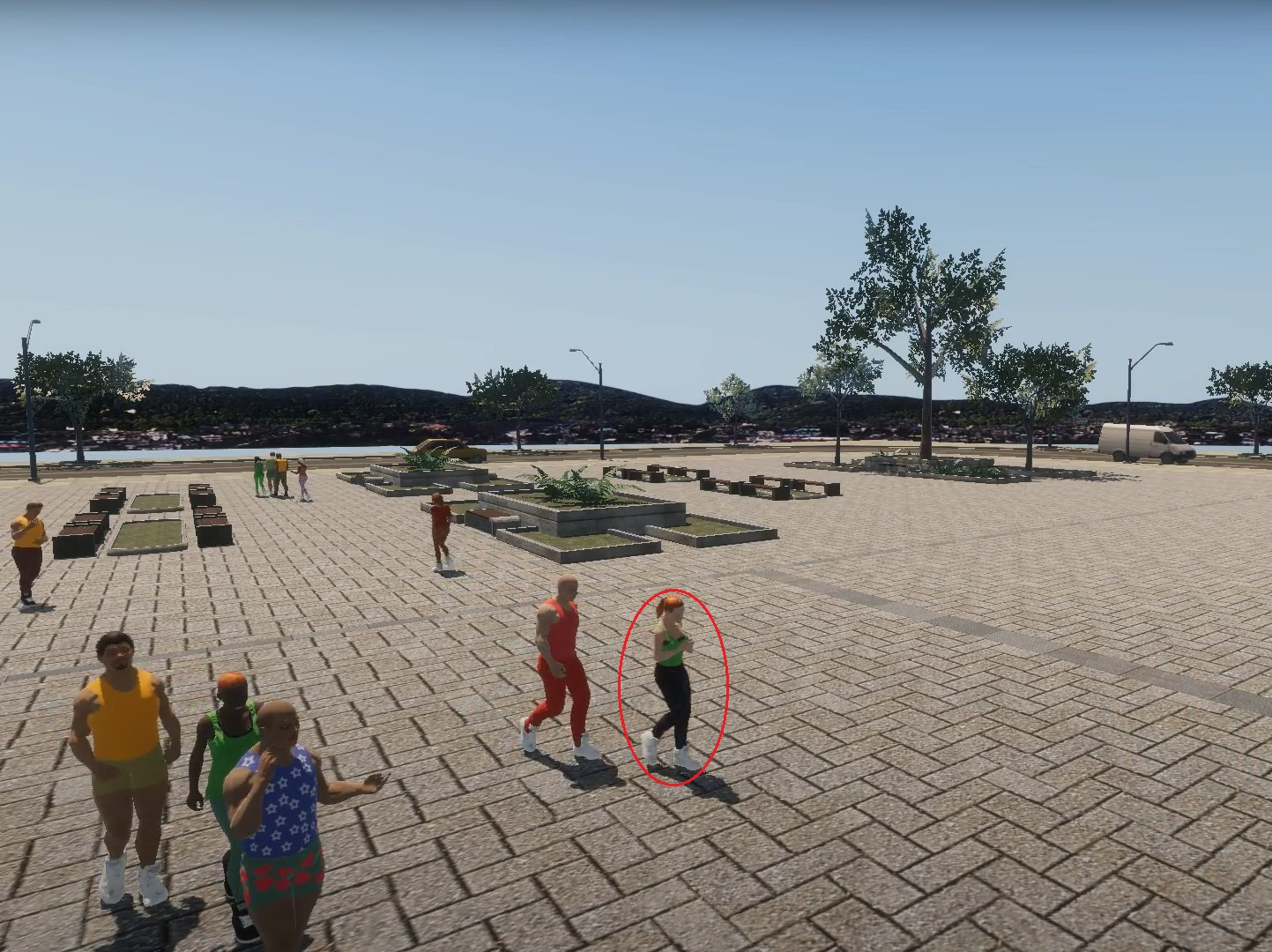}}
    \subfigure[]{\includegraphics[height=0.165\textwidth]{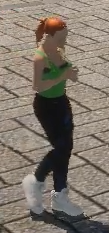}}
    \subfigure[]{\includegraphics[height=0.165\textwidth]{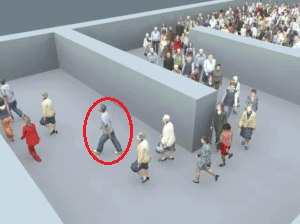}}
    \subfigure[]{\includegraphics[height=0.165\textwidth]{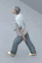}}
    \subfigure[]{\includegraphics[height=0.165\textwidth]{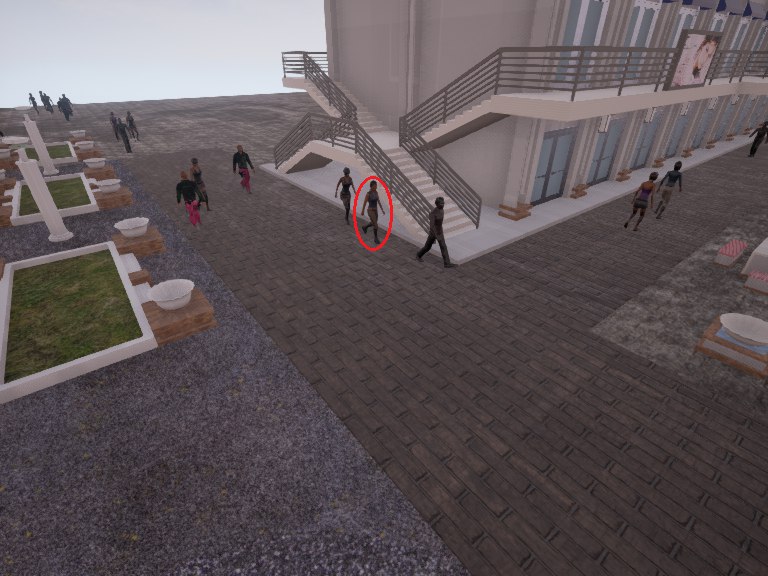}}
    \subfigure[]{\includegraphics[height=0.165\textwidth]{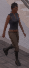}}
    \caption{Visual comparison of the proposed (a,b) simulation framework against (c, d) Agoraset \cite{allain2012agoraset}, and (e, f) LcrowdV \cite{cheung2016lcrowdv} datasets. Agoraset lacks realism; LcrowdV has low resolution, limited ground truth data for different tasks, and low-quality rendering (e.g. light simulation, shadows).}
    \label{fig:visual_comparison}
\end{figure*}

Available crowd simulators \cite{allain2012agoraset,cheung2016lcrowdv, curtis2016menge}, mostly focus on the behavioral simulation, sometimes totally omitting the rendering of the visuals \cite{curtis2016menge}. As shown in Fig. \ref{fig:visual_comparison}, even when the rendering is provided, the visual fidelity and the scene appearance are of poor quality, which limits its use in computer vision.

\begin{table*}[]
\centering
\resizebox{\textwidth}{!}{%
\begin{tabular}{@{}cccccccc|cc@{}}
\toprule
\multirow{2}{*}{\textbf{Dataset}} & \multicolumn{7}{c|}{\textbf{Ground Truth}}                                                 & \multirow{2}{*}{\textbf{\begin{tabular}[c]{@{}c@{}}Behavioral \\ fidelity\end{tabular}}} & \multirow{2}{*}{\textbf{\begin{tabular}[c]{@{}c@{}}Visual \\ fidelity\end{tabular}}} \\ \cmidrule(lr){2-8}
                                  & Segmentation & Detection  & Tracking   & Trajectory & Anomaly    & HPE        & Counting   &                                                                                          &                                                                                      \\ \midrule
JTA \cite{fabbri2018learning}            & \faTimes            & \faTimes          & \faCheck          & \faTimes          & \faTimes          & \faCheck          & \faTimes          & \faTimes                                                                                        & \faCheck                                                                                    \\
MothSynth \cite{fabbri21iccv}     & \faCheck            & \faTimes          & \faCheck          & \faTimes          & \faTimes          & \faCheck          & \faTimes          & \faTimes                                                                                        & \faCheck                                                                                    \\
Viper \cite{richter2017playing}              & \faCheck            & \faTimes          & \faTimes          & \faTimes          & \faTimes          & \faCheck          & \faTimes          & \faTimes                                                                                        & \faCheck                                                                                    \\
GTA \cite{krahenbuhl2018free}            & \faCheck            & \faTimes          & \faTimes          & \faTimes          & \faTimes          & \faTimes          & \faTimes          & \faTimes                                                                                        & \faCheck                                                                                    \\
PFD \cite{richter2016playing}            & \faCheck            & \faTimes          & \faTimes          & \faTimes          & \faTimes          & \faTimes          & \faTimes          & \faTimes                                                                                        & \faCheck                                                                                    \\
Synthia \cite{ros2016synthia}            & \faCheck            & \faTimes          & \faTimes          & \faTimes          & \faTimes          & \faTimes          & \faTimes          & \faTimes                                                                                        & \faCheck                                                                                    \\\midrule
Menge \cite{curtis2016menge}            & \faCheck            & \faTimes          & \faTimes          & \faTimes          & \faCheck          & \faCheck          & \faTimes          & \faCheck                                                                                        & \faTimes                                                                                    \\
AgoraSet \cite{allain2012agoraset}       & \faTimes            & \faCheck          & \faTimes          & \faCheck          & \faTimes          & \faTimes          & \faCheck          & \faCheck                                                                                        & \faCheck                                                                                    \\
LCrowdV \cite{cheung2016lcrowdv}         & \faTimes            & \faCheck          & \faTimes          & \faCheck          & \faTimes          & \faTimes          & \faCheck          & \faCheck                                                                                        & \faCheck                                                                                    \\
\textbf{Ours}                     & \faCheck   & \faCheck & \faCheck & \faCheck & \faCheck & \faCheck & \faCheck & \faCheck                                                                               & \faCheck                                                                           \\ \bottomrule
\end{tabular}%
}
\vspace{5px}
\caption{Synthetic datasets comparisons; a complete overview can be found at \cite{nikolenko2021synthetic}. The first six methods provide synthetic datasets for deep learning applications, which are generated without validating the behavioral fidelity of the involved agents. The last three methods also provide full simulation engines, which can be used by researchers to generate new data. They also validate a behavioral model. As can be seen, our proposal provides the most complete setup.}
\label{tab:simcomparison}
\end{table*}

\section{Related work}
\label{sec:soa_review}

\textbf{Image and video-based labeled data: application areas and shortcomings.} 
The current evolution of machine and deep learning in image processing and computer vision has shown countless applications including, among others, object localization \cite{zhang2021weakly} human action recognition \cite{zhang2019comprehensive}, human pose estimation \cite{kanazawa2018end,cao2019openpose}, and 3D reconstruction \cite{chang2015shapenet}. Each of them has brought up, in the relevant research community, the need for bigger datasets, with improved diversity, in terms of representation of the environment, camera modeling and pose, and lighting. Among the most common and general-purpose datasets in the literature, MS COCO (COmmon Object in Context) \cite{lin2014microsoft}, PASCAL VOC \cite{everingham2010pascal} and KITTI \cite{geiger2012we} provide an excellent resource for many application scenarios.
When it comes to the analysis of human-related information, Human 3.6 \cite{ionescu2013human3} and Panoptic \cite{Joo_2017_TPAMI} are, as of today, the most exhaustive datasets, providing multiple viewpoints of the human body along with the corresponding ground truth. 

In other scenarios, like in monitoring and surveillance, the availability of relevant data is scarce, as the footage exhibiting events of interest is generally limited, and poorly annotated. 
When dealing with people detection and tracking \cite{sheng2020hypothesis}, action recognition \cite{li2019recurrent}, behavior analysis \cite{kang2018beyond}, counting \cite{sajid2020zoomcount}, a lot of complications arise as the scene becomes crowded. 
In such cases, also basic tasks as counting people come at a high cost in terms of annotation, generally leading to questionable quality when the resolution of the subjects in the picture becomes too small. 

The same applies to the analysis of human behavioral patterns. As an example, in the human trajectory prediction domain, researchers have been relying on the UCY \cite{lerner2007crowds} and ETH \cite{pellegrini2009you} datasets for a long time. These datasets consist of around 400 trajectories in total, lasting a few seconds each. Given the small size of the datasets, researchers \cite{alahi2016social} have been using different workarounds, such as generating data according to the Social Forces Model (SFM) \cite{helbing1995social} to achieve better performances. 
Other datasets like the Stanford drone dataset \cite{robicquet2020learning} provide a bigger number of trajectories, though from a completely different perspective, and in a different context (skateboarders, bikers), neither allowing for knowledge transfer nor domain adaptation. 

\textbf{Synthetic datasets and simulation in computer vision.} The use of synthetic datasets in computer vision is not new per se \cite{Butler:ECCV:2012,mayer2018makes,ros2016synthia,gaidon2016virtual,baslamisli2018joint},
and the recent advancements in the video-games industry have enabled the development of improved and highly desirable graphical representations. Among the early contributions is the MPI-Sintel dataset \cite{Butler:ECCV:2012}, meant for optical flow analysis, which has been widely used as ground-truth source for depth estimation and bottom-up segmentation.

Many applications like autonomous and aerial systems, have a big interest in exploring the field of simulation \cite{li2017paralleleye,chen2015deepdriving,dosovitskiy2017carla,shah2018airsim}, mostly because the data collection in the real world is expensive and potentially harmful in case of accidents. 

Through simulation, it is possible to create large datasets consisting of images, videos, metadata, comprising of accurate and automatically-generated ground truth, in the form of bounding boxes, per-pixel depth, optical flow, semantic classes and instance segmentation, exploiting video games-like paradigms \cite{richter2016playing,richter2017playing,savva2017minos}.

An exhaustive overview on the use of synthetic datasets in image processing and computer vision is reported in \cite{mayer2018makes}.

The use of computer-generated data has motivated researchers in investigating the effectiveness of the synthetic data \cite{gaidon2018reasonable,johnson2016driving,savva2017minos,gaidon2016virtual}, and how much strategic they are in solving real-world problems. For this reason procedural generation of synthetic videos \cite{de2017procedural} as well as the integration of 3D engines \cite{qiu2016unrealcv} have been widely exploited, and the available approaches aim at creating tools capable on the one hand to generate realistic scenes, and on the other hand produce the related ground truth.
More recent approaches have employed modern video-game rendering to obtain the best possible appearance \cite{fabbri2018learning,fabbri21iccv}.
In simulation, other than macro behavioral and visual fidelity, it is also relevant to provide realistic animations of people's movements. Humans are capable of discerning between plausible or artificial poses and movements, even when the synthetic visual fidelity level is on par with the real world fidelity. A solution to the issue of artificial humanoid movements is to transpose the movements of real people to synthetic agents via motion capture frameworks or human pose estimation.
In the crowd analysis domain, Agoraset \cite{allain2012agoraset} and LcrowdV \cite{cheung2016lcrowdv} have introduced visual simulations to create synthetic datasets. LcrowdV \cite{cheung2016lcrowdv} focuses on validating the \textit{visual fidelity} on people detection. Agoraset \cite{allain2012agoraset} focuses on validating the \textit{behavioral fidelity}, providing validation for the generated trajectories. In our work, we tackle both aspects, aiming at consolidating both the visual appearance features and the behavioral modules of crowd simulators.

\section{The proposed simulator}
\label{sec:casestudy}

Crowd analysis and monitoring has been largely investigated as a mean to improve the safety of people \cite{allain2012agoraset,de2016detection}. 
The main limitation when applying visual machine learning algorithms to crowded scenes is the need for an accurate ground truth reporting the pedestrians' position: the annotations, however, tend to be unreliable when dealing with very crowded scenarios (e.g. crowd counting at a concert), because of mutual occlusions and limited size of the subjects involved.
To cope with these problems, the adoption of simulators and synthetic data generation have been recently explored. 
While the existing literature in computer vision has used synthetic data to address the need of very specific tasks (e.g. detection \cite{fabbri21iccv}), we introduce UniCrowd, a crowd simulator that can generate synthetic datasets of arbitrary size and complexity. We show how our simulator is unique, as it can produce data, which can tackle different challenges in a multi-modal fashion.

In fact, most real-world datasets are often conceived for specific tasks, and are provided with ad-hoc ground-truth labels. Leveraging our simulator, we can provide ground truth that is compatible with multiple tasks both on the behavioral side, such as trajectory prediction, and anomaly detection, as well as on the appearance side, like people detection and segmentation, as well as crowd counting, human pose estimation, and anomaly detection.

UniCrowd is meant to meet the requirements for both \textit{behavioral fidelity} and \textit{visual fidelity}.
On the \textit{behavioral fidelity} side, a crowd simulator has to manage the crowd movements on macro and micro perspectives. The compliance with macro crowd behavior consists in reproducing patterns, which are typical of the crowd as a whole, such as the emergent behavior of people going to the same direction, forming lines, or the crowd following social rules, such as walking along a pathway \cite{de2016detection}. The macro rules can change depending on cultural factors, such as the perception of the personal space, which can be different across different continents \cite{hall1968proxemics}. 
On the other hand, micro crowd behaviors focus on the individual, dealing with the avoidance of obstacles and other people in the crowd; this involves the personal sphere (e.g. shyness, aggressiveness \cite{cheung2016lcrowdv}) and it is driven by the current circumstances (e.g. being in a hurry to catch a bus).

As for the \textit{visual fidelity}, representing a crowd of humans means dealing with appearance and motion features that closely match what a human observer would see in the real life, modeling the environment around the crowd, introducing weather and light changes throughout the day. 
Ideally, synthetic data should be as photo-realistic and close to the real world as possible. In case of RGB data, that means working with the fine details of light and object shaping, a very costly feature that only top-tier video-games can afford. 
However, video-games usually lack a sufficient \textit{behavioral fidelity}, with pedestrians and vehicles in the scenes going through predefined paths with little behavioral realism, and poor customization options.

\begin{figure*}
    \centering
    \includegraphics[width=.994444\textwidth]{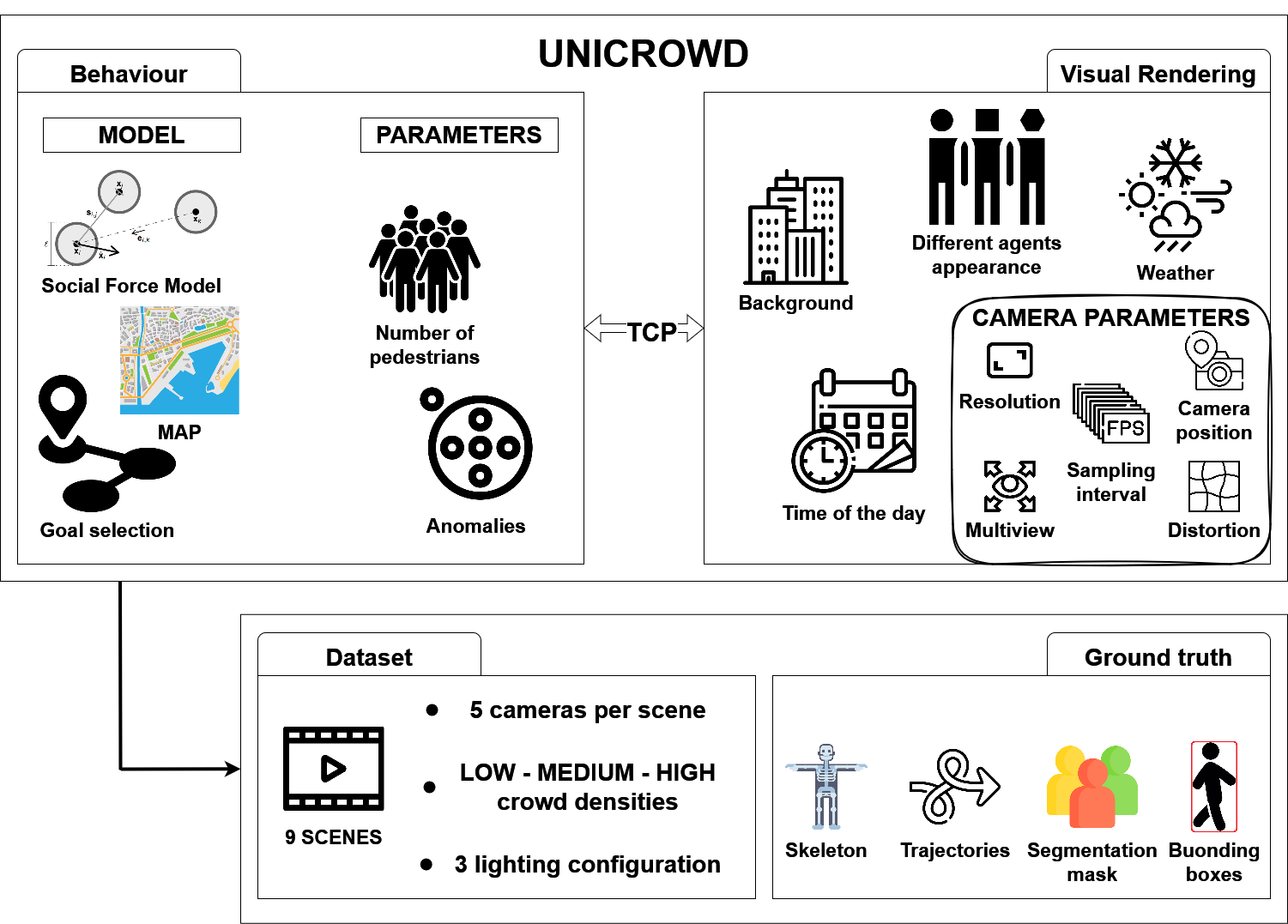}
    \caption{Our simulator is structured with two blocks, one implementing the behavioral model and the other one providing the visual engine for graphical representation. The two modules communicate through a TCP socket. They can be individually replaced or updated, in terms of behavioral models or visual appearance of the simulation. The output of the simulator consists of the video sequence, together with the corresponding set of ground truths.}
    \label{fig:teaser_sim}
\end{figure*}

\begin{figure*} 
\centering 
  \begin{minipage}[b]{0.45\linewidth}
    \centering
    \includegraphics[width=\textwidth]{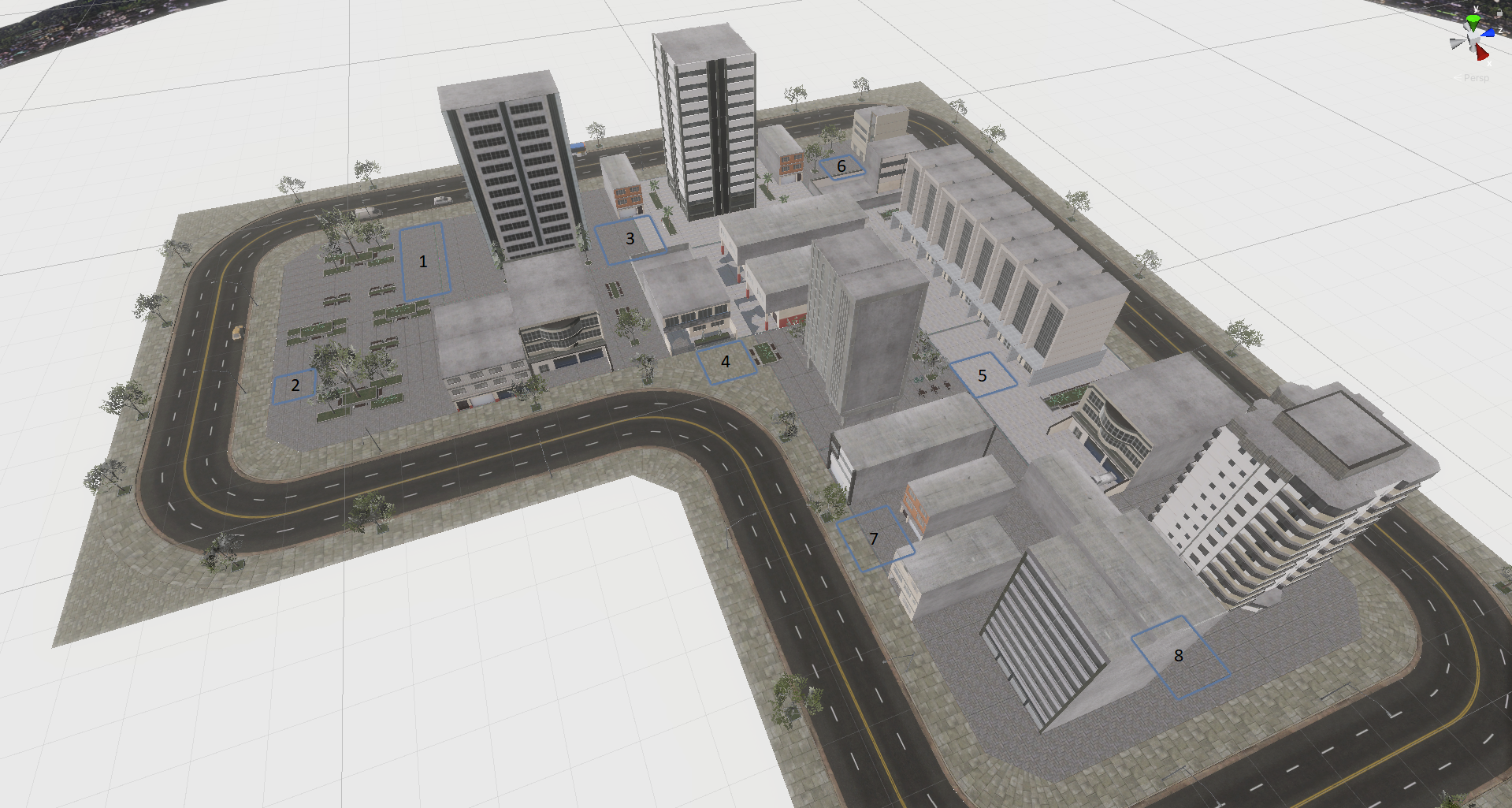}
    \caption{Synthetic human agents can spawn from different generation areas in the environment.}
    \label{fig:generation_areas}
    \vspace{2ex}
  \end{minipage}
  \hfill
  \begin{minipage}[b]{0.45\linewidth}
    \centering
    \includegraphics[width=\textwidth]{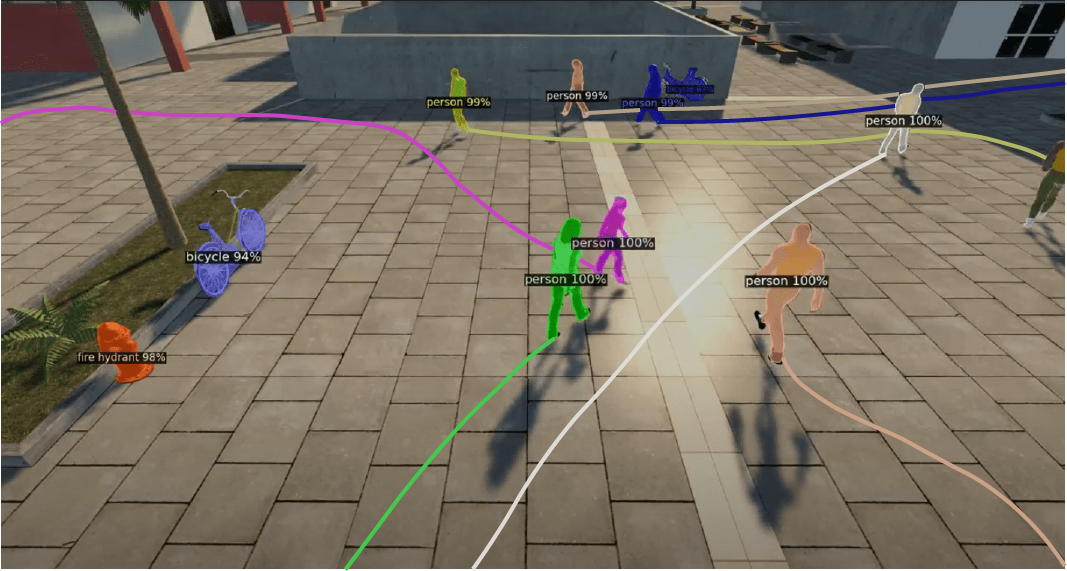}
    \caption{Precise human trajectories can be easily computed without manual annotation inside the simulated environment.}
    \label{fig:trajectories}
  \end{minipage} 

  \begin{minipage}[b]{0.45\linewidth}
    \centering
    \includegraphics[width=\textwidth]{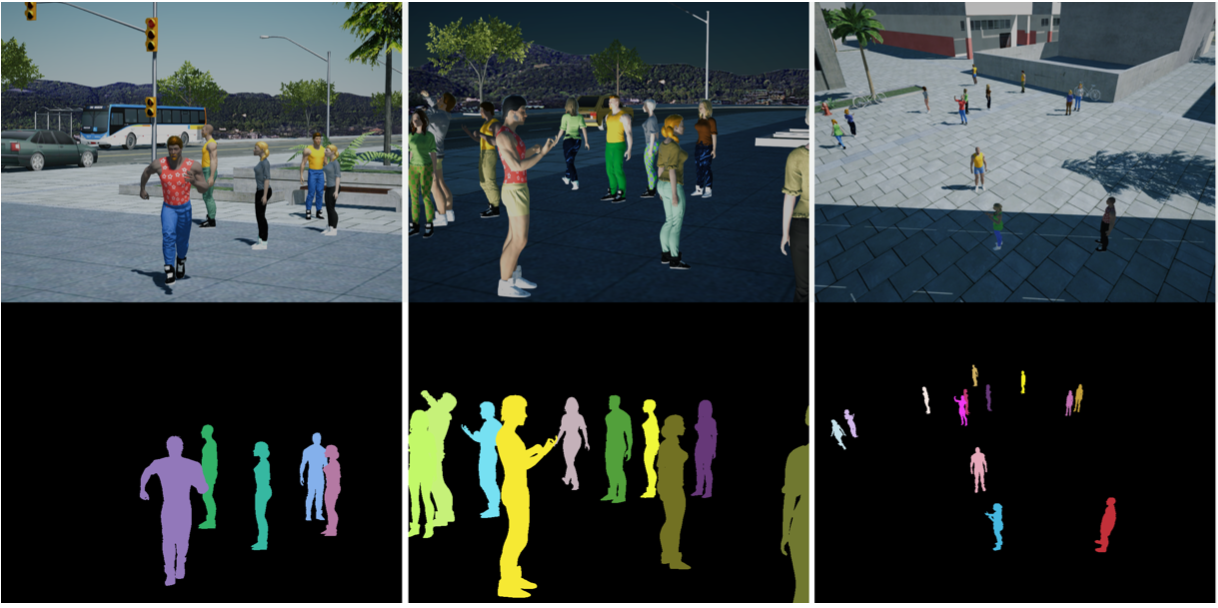}
    \caption{Samples from our training dataset, with associated instance segmentation mask.}
    \label{fig:segmentation_data}
  \end{minipage}
  \hfill
\begin{minipage}[b]{0.45\linewidth}
    \centering
    \includegraphics[width=\textwidth]{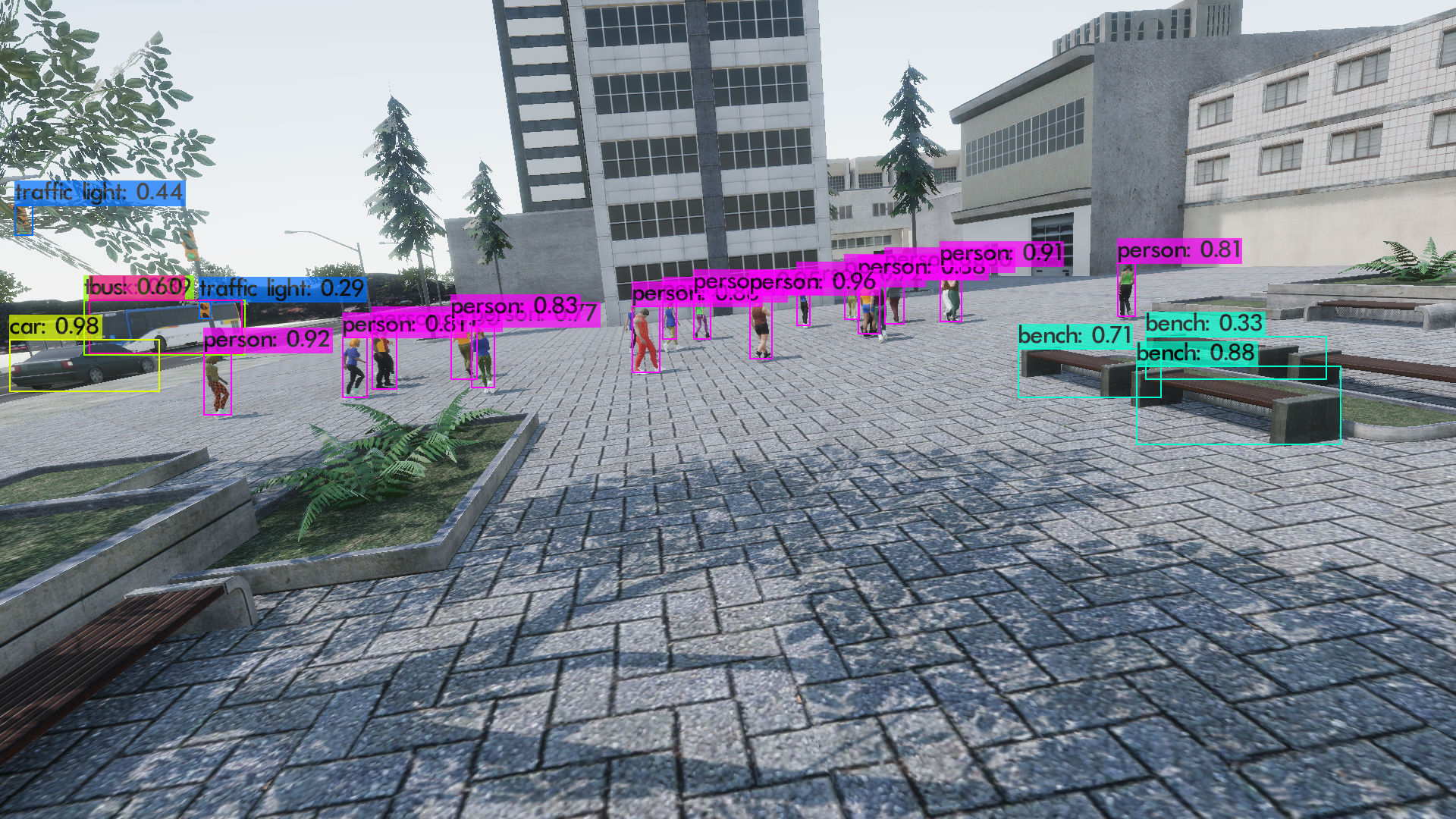}
    \caption{Common object detection algorithms \cite{yolov3} display accurate performances in the simulated environment.}
    \label{fig:yolo_on_sim} 
  \end{minipage}

  \begin{minipage}[b]{0.45\linewidth}
    \centering
    \includegraphics[width=\textwidth]{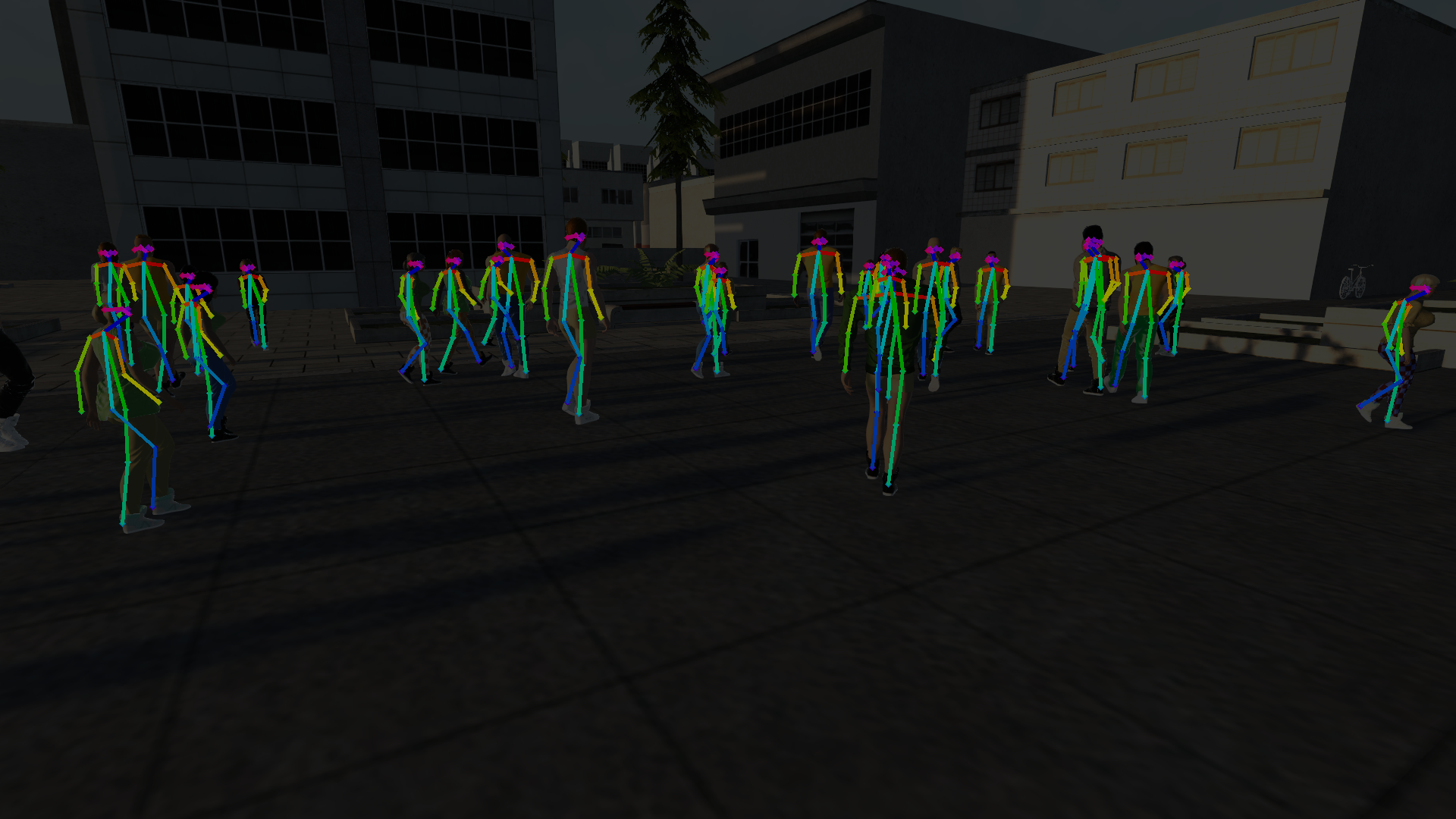}
    \caption{Precise ground truth joint annotations for human pose estimation can be saved at each timestamp during simulation.}
    \label{fig:pose1}
    \vspace{4ex}
  \end{minipage}
  \hfill
    \begin{minipage}[b]{0.45\linewidth}
    \centering
    \includegraphics[width=\textwidth]{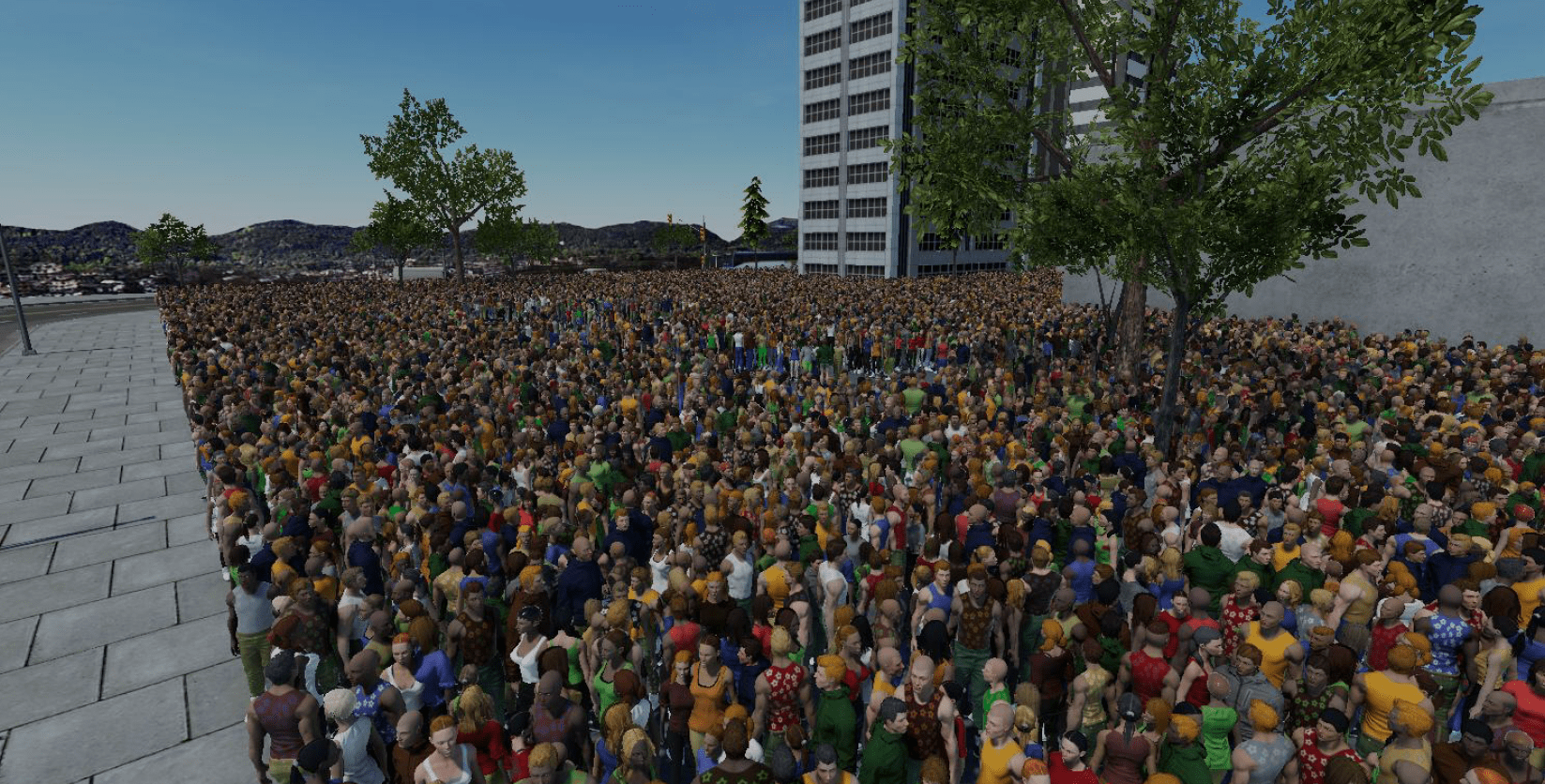}
    \caption{Each simulation can be run with an arbitrary number of agents, providing a ground truth suitable also for crowd counting tasks.}
    \label{fig:counting_sim}
    \vspace{4ex}
  \end{minipage} 

\end{figure*}

\begin{figure*}
    \centering
    \subfigure[]{\includegraphics[width=0.2\textwidth]{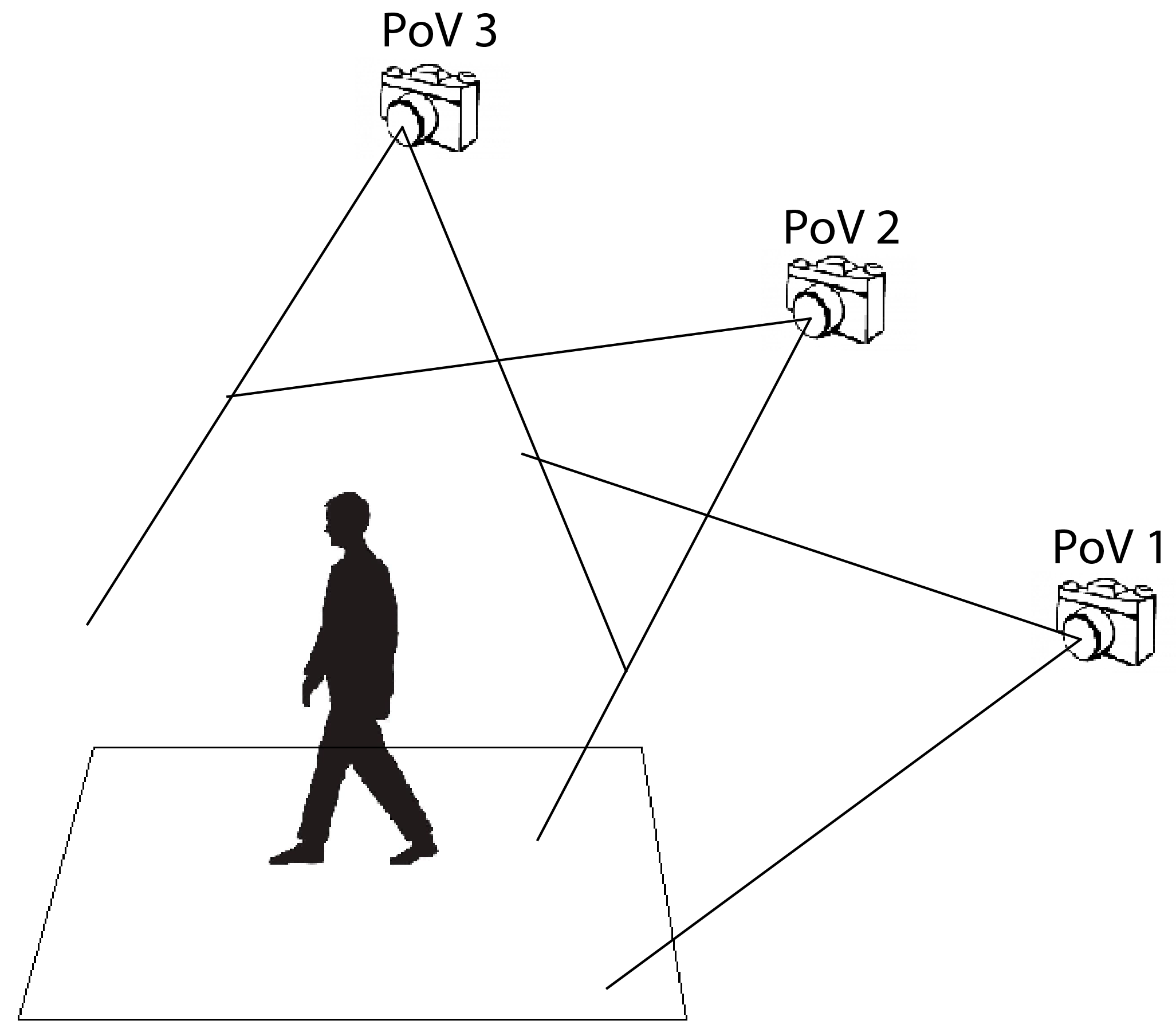}}
    \subfigure[]{\includegraphics[width=0.26\textwidth]{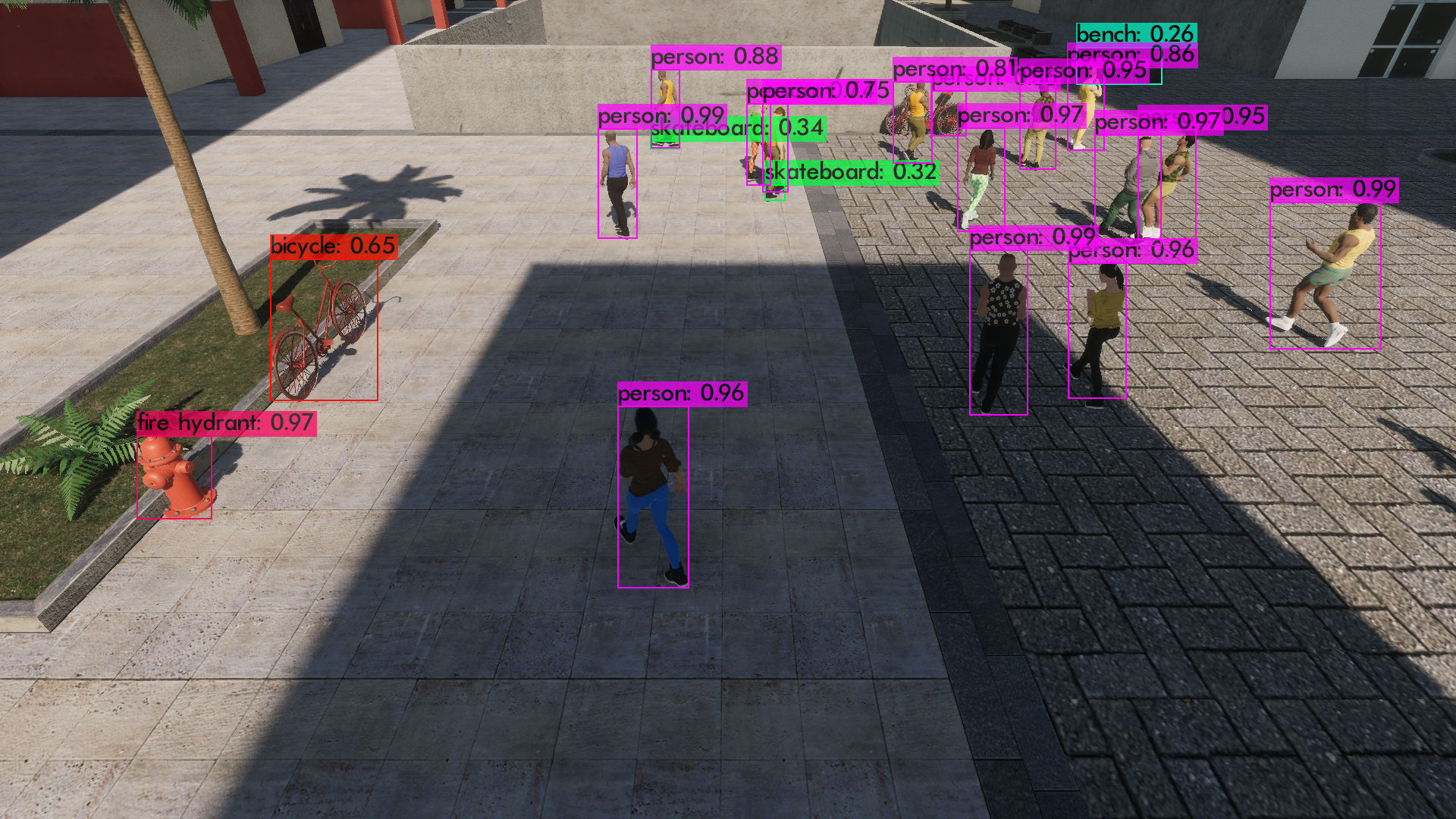}}
    \subfigure[]{\includegraphics[width=0.26\textwidth]{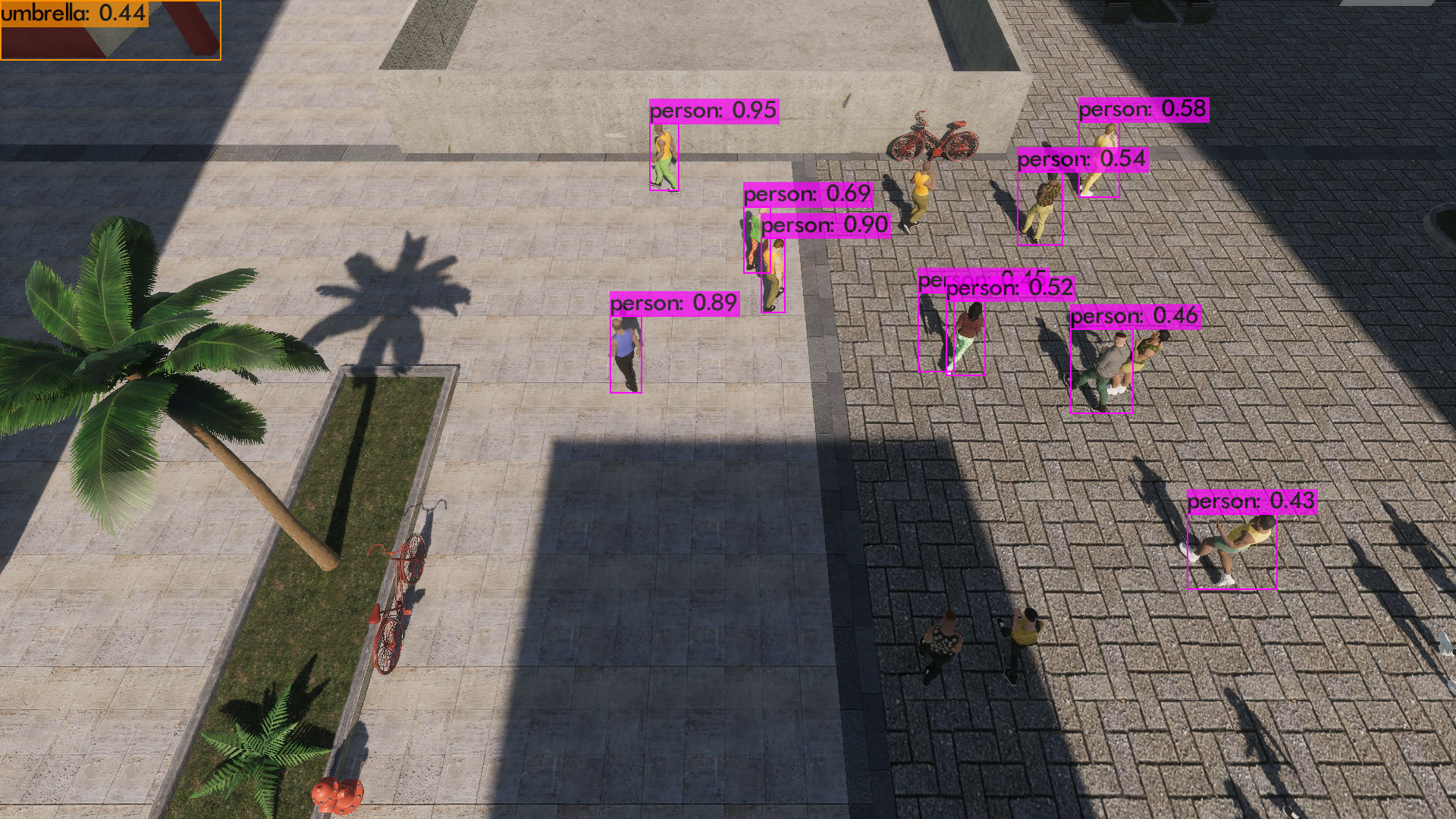}}
    \subfigure[]{\includegraphics[width=0.26\textwidth]{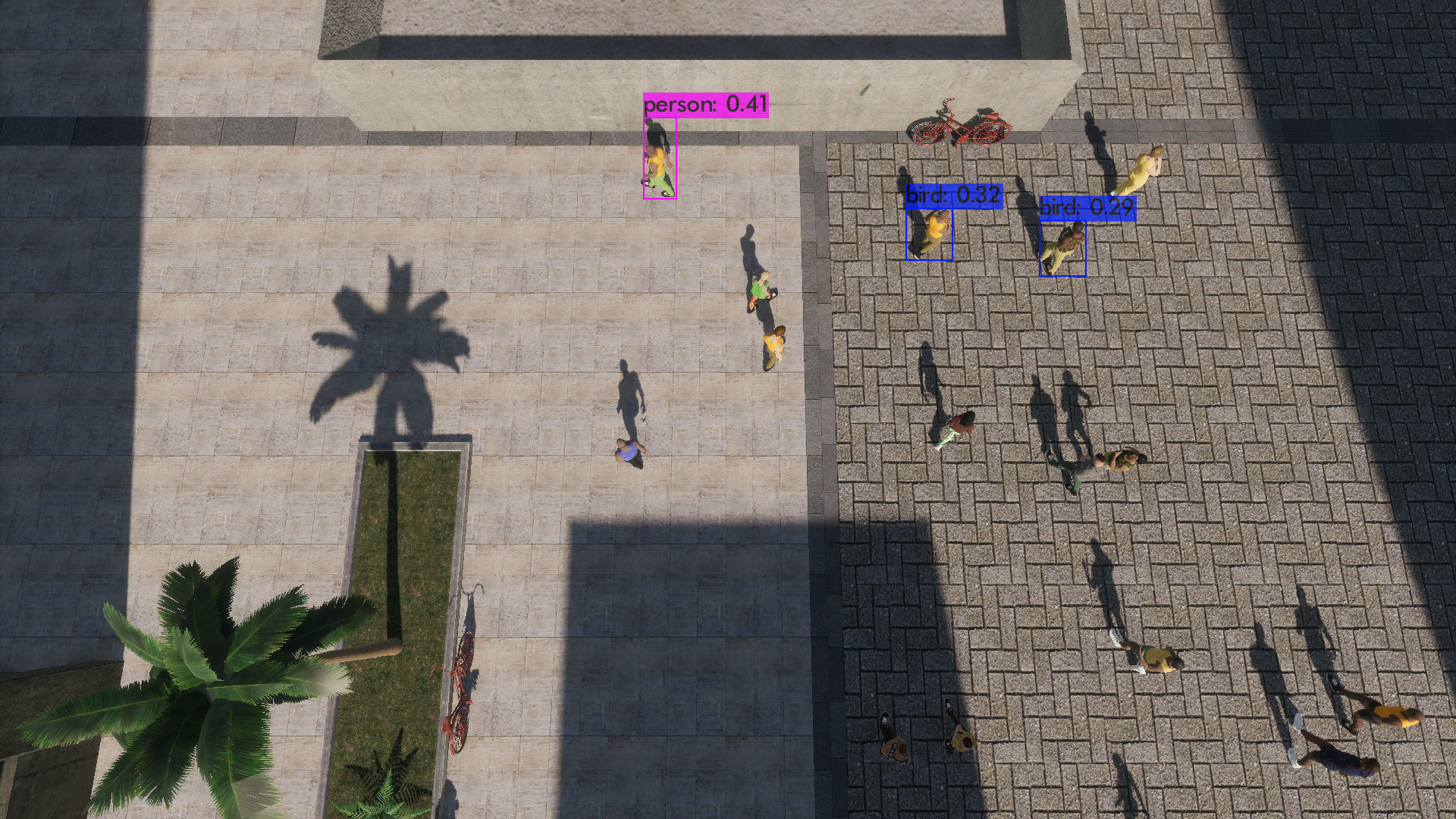}}
    \caption{Varying the camera viewpoint (a) causes changes in the performances of common detection algorithms. Using the simulated environment we can provide training data from arbitrary views (b, c, d), enriching the datasets to tackle challenging scenarios, like the top-viewpoint (d), in which standard detectors perform poorly.}
    \label{fig:viewpoints}
\end{figure*}

\begin{figure*}[!ht]
    \centering
    \resizebox{\textwidth}{!}{
    \begin{tabular}{c|ccc}
          & \textbf{7.00} & \textbf{12.00} & \textbf{18.30} \\
    \midrule
        \rotatebox[origin=c]{90}{\textbf{Detectron}} & 
        \raisebox{-0.5\height}{\includegraphics[width=.28\textwidth]{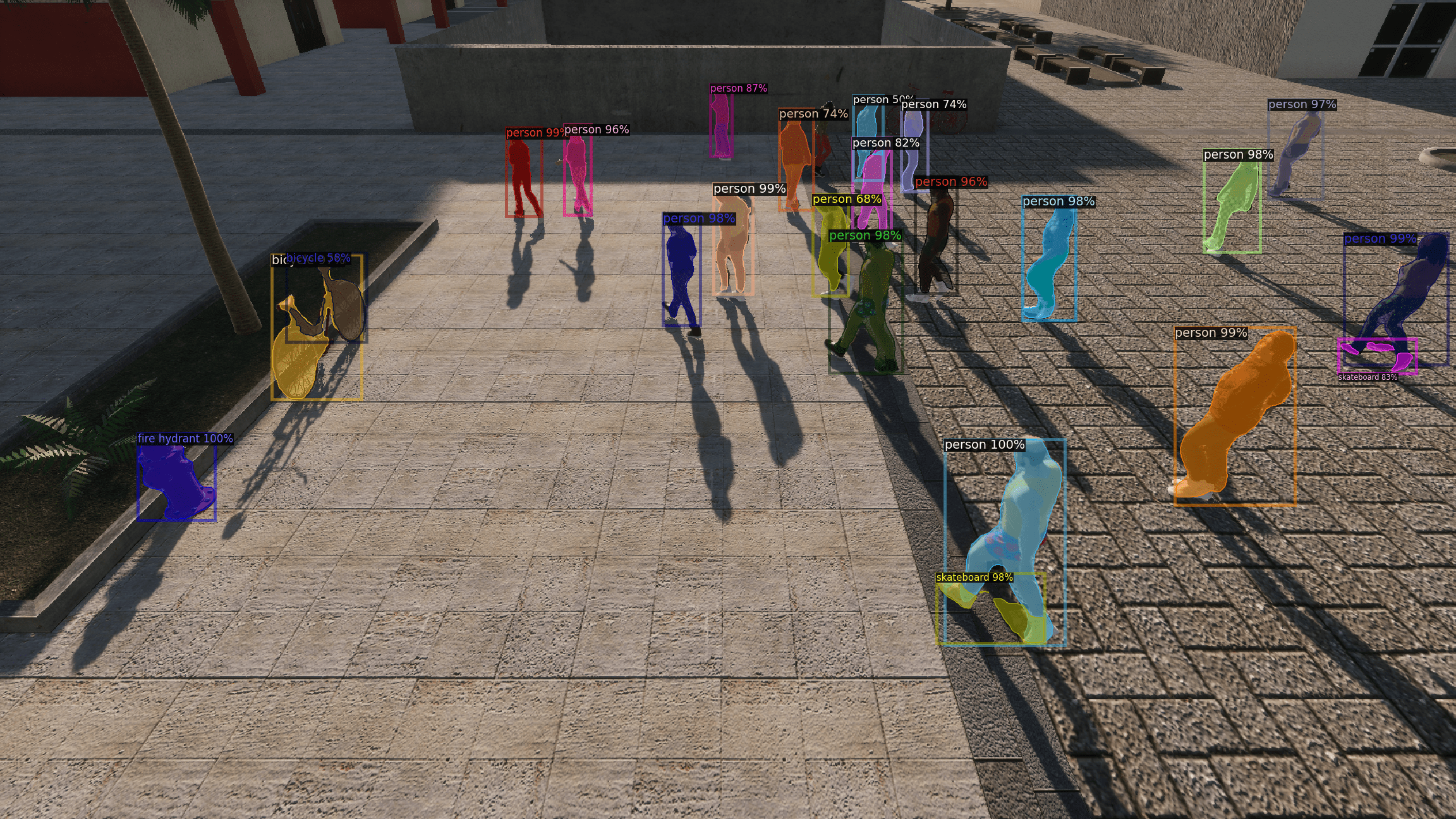}} &
        \raisebox{-0.5\height}{\includegraphics[width=.28\textwidth]{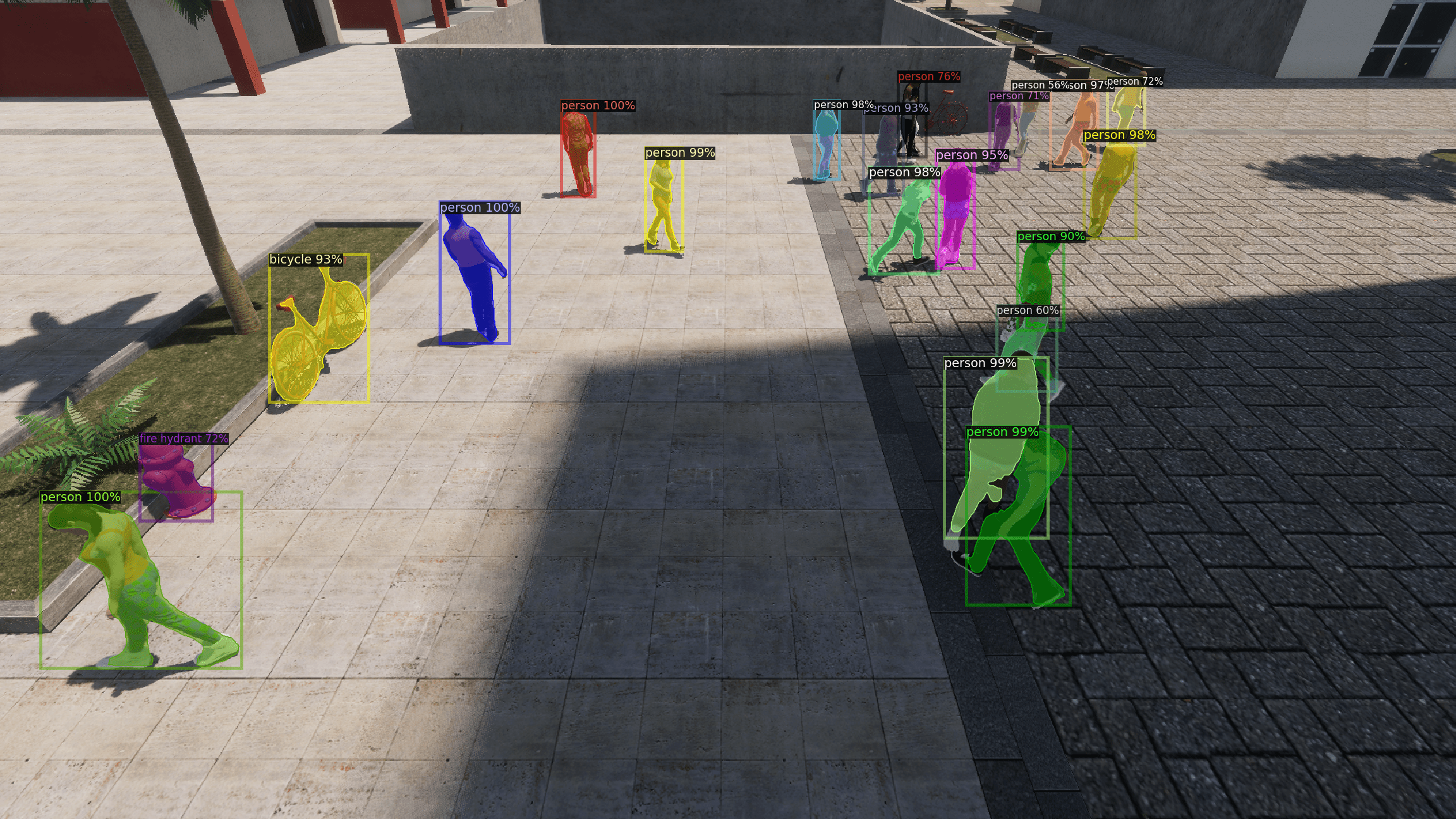}} & \raisebox{-0.5\height}{\includegraphics[width=.28\textwidth]{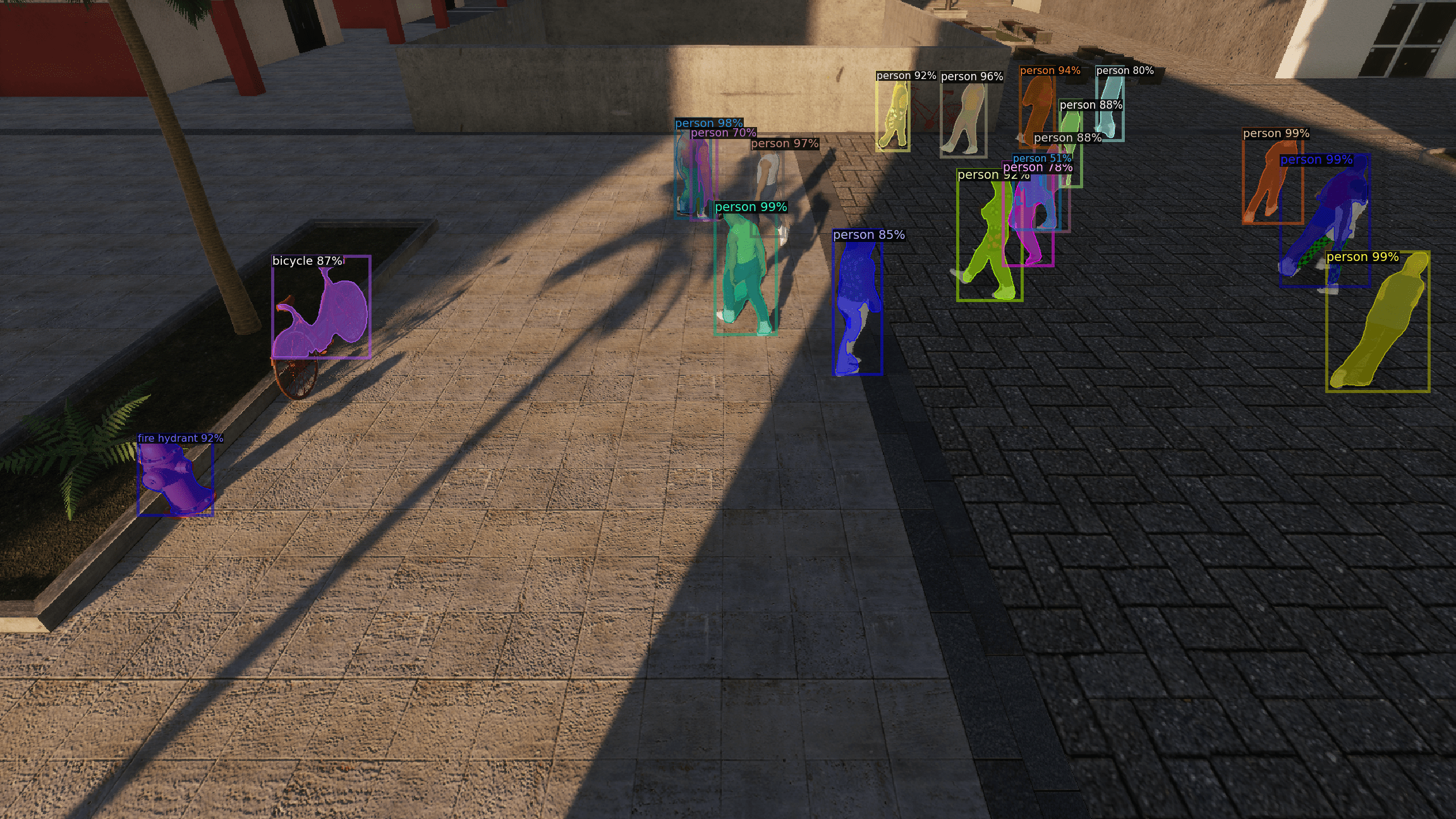}} \\
    \midrule
        \rotatebox[origin=c]{90}{\textbf{YOLO}} & 
        \raisebox{-0.5\height}{\includegraphics[width=.28\textwidth]{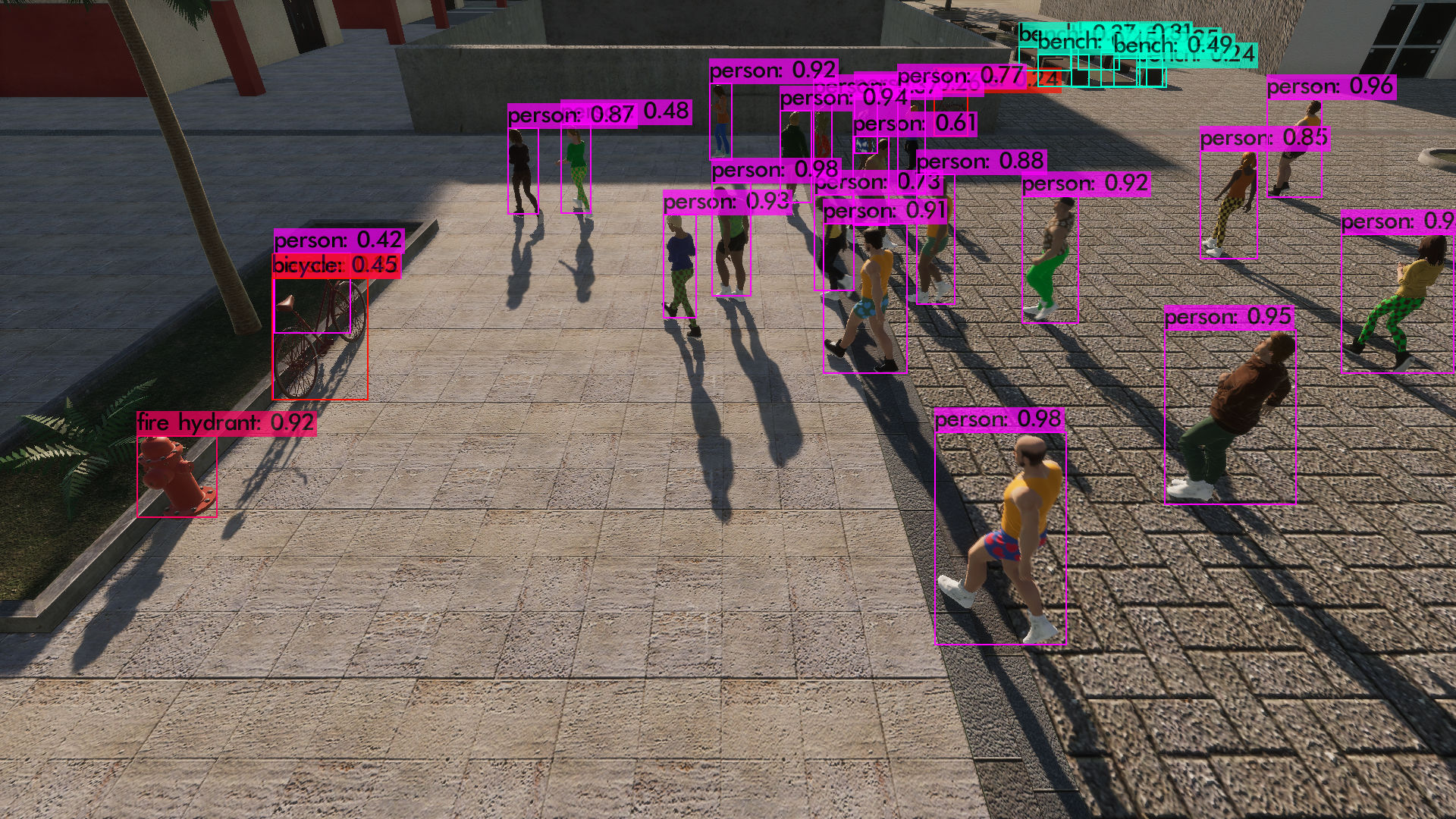}} & \raisebox{-0.5\height}{\includegraphics[width=.28\textwidth]{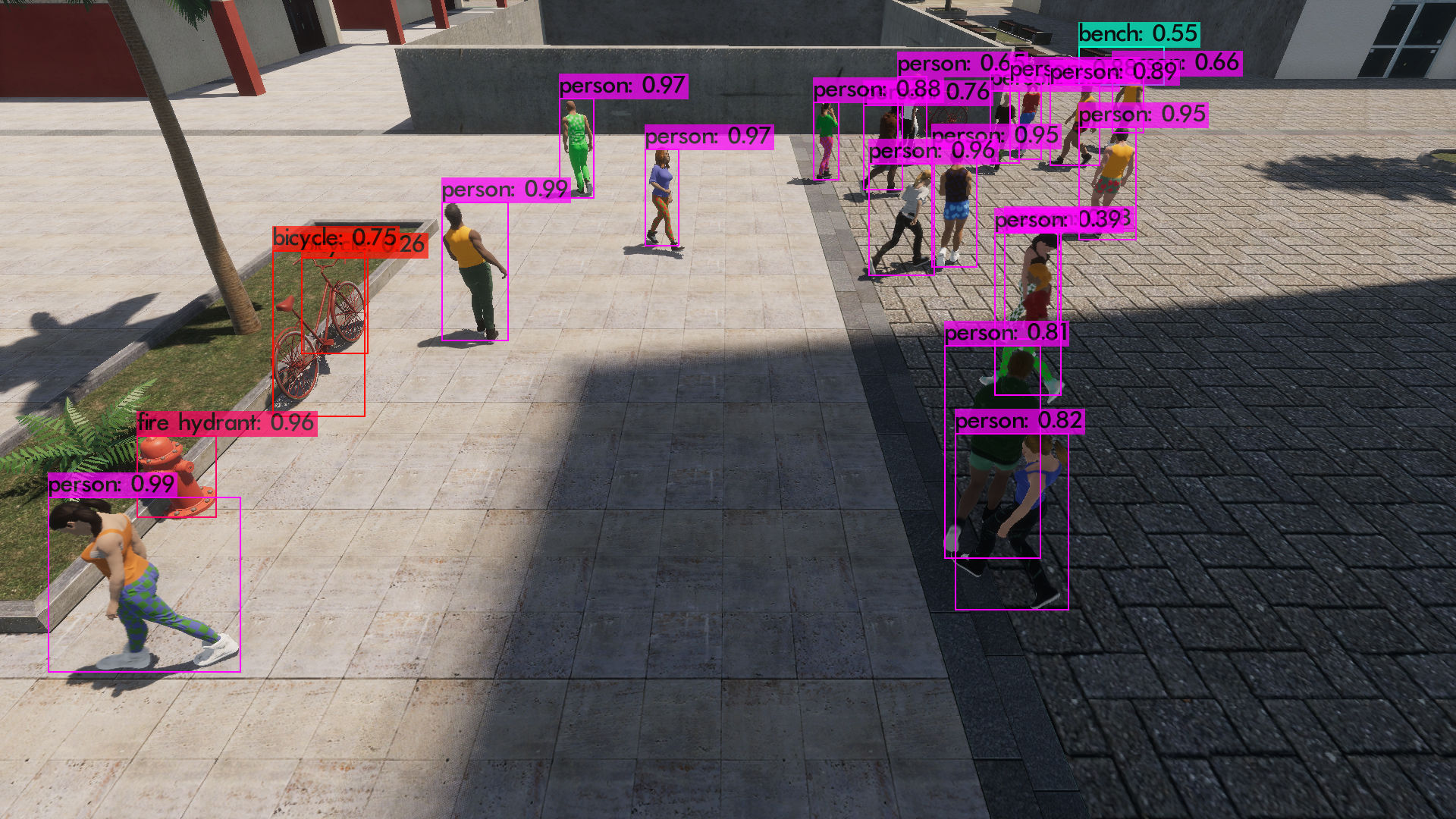}} & \raisebox{-0.5\height}{\includegraphics[width=.28\textwidth]{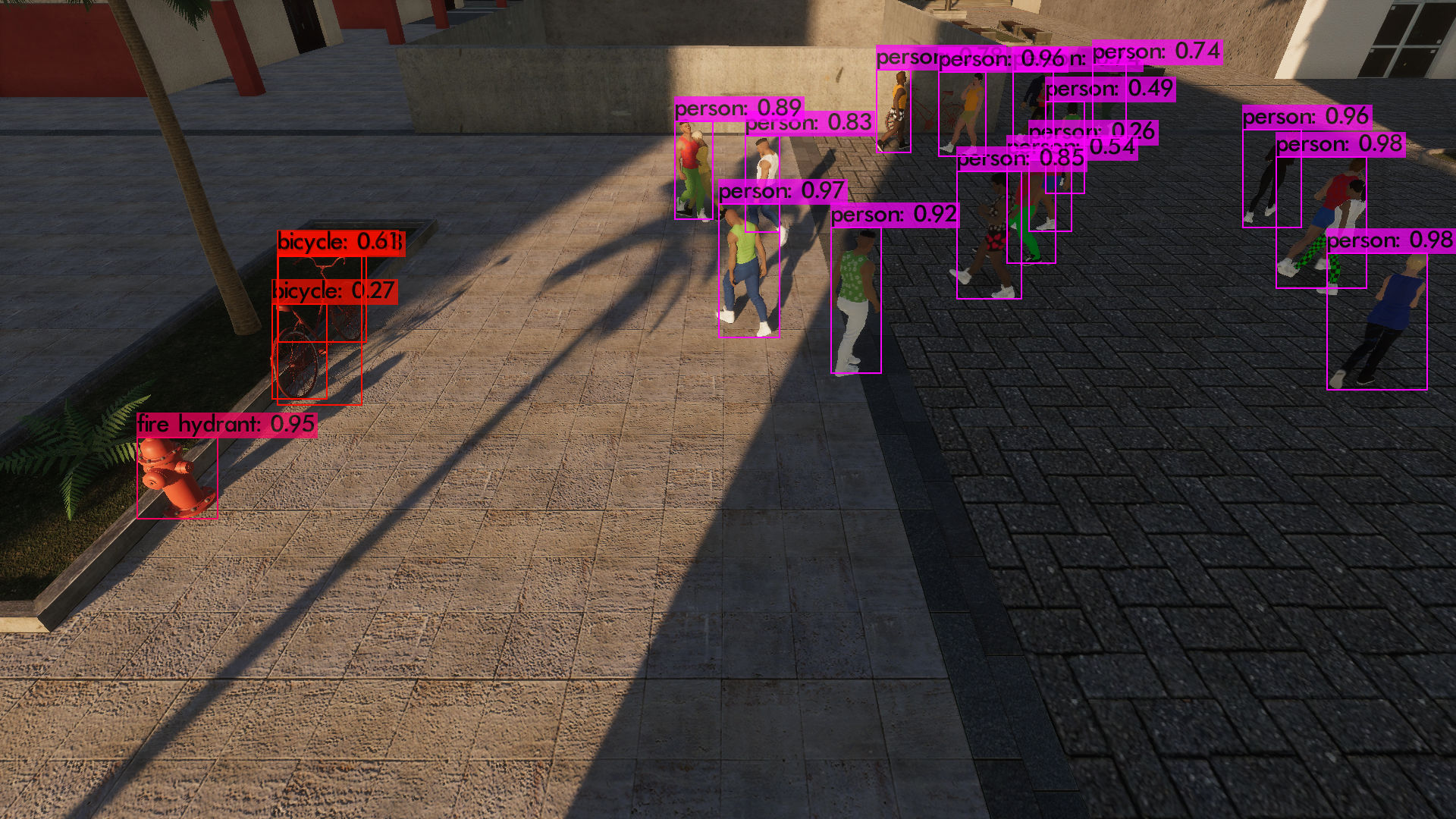}} \\
    \end{tabular}
    }
    
    \caption{Detectron vs YOLO: detection performances on the synthetic data at different times of the day for a given view.}
    \label{fig:segm_results}
\end{figure*}

\subsection{Architecture and components}

To meet the aforementioned requirements for both visual and behavioral fidelity, we Unicrowd is be able to independently develop the behavioral and appearance components, as shown in Fig. \ref{fig:teaser_sim}. 

\subsubsection{Behavioral module}

On the behavioral side, we deploy a model based on the Social Forces Model (SFM) \cite{helbing1995social}. The SFM is an agent-based model, where each agent computes its next position given its current velocity, and repulsive forces with respect to obstacles and other pedestrians. SFM has proven to effectively model the micro behavior of the crowd, while retaining a good approximation of the macroscopic behaviors. Compared to other solutions \cite{van2008reciprocal,treuille2006continuum}, which well reproduce the crowd behavior in well defined contexts, the SFM has proven its ability to better generalise to a multitude of scenarios. Moreover, it can be refined and enriched to display more sophisticated behaviors \cite{moussaid2010walking,Yu2005}.

The behavioral section of the simulator is decoupled from the graphical part: given the agent's position at time step $t$, and considering its final destination and other personal parameters, it produces the next position at time step $t+1$. The position is then fed to the graphical rendering block using a TCP socket. Differently from other simulators where the scenario is predefined off-line, the TCP socket enables the two modules to dynamically exchange information at run-time. Thus, a change in the environment at the graphical level is directly communicated to the behavioral module allowing for dynamic adaptation.
Having the two macro-blocks separated opens to the possibility of controlling the simulation in a dynamic fashion, generating events and anomalies, instead of being bounded to pre-computed simulation scripts like in other competing solutions \cite{curtis2016menge,cheung2016lcrowdv,allain2012agoraset}.

As shown in Fig. \ref{fig:teaser_sim}, the user is given control to a number of parameters, such as:
\begin{itemize}
    \item number of pedestrians involved in the simulation, or the desired crowd density.
    \item pedestrian's behavior modelling parameters, such as preferred velocity, anomalous behaviors and preferred social space.
    \item spawn and goals point for each pedestrian (Fig. \ref{fig:generation_areas}).
    \item environment map and walkable areas.
\end{itemize}

\subsubsection{Visual rendering}

\begin{figure}[!ht]
\centering
\includegraphics[width=0.5\textwidth]{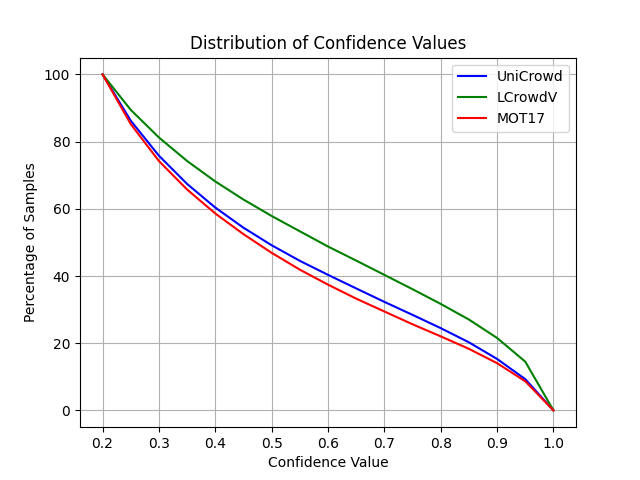}
\caption{To evaluate the visual fidelity of the proposed simulator, we run an off the shelf version of YOLO on our dataset, LcrowdV, and, for comparison, the video sequences from the MOT17 challenge. In this graph, we show the curve of the percentages of bounding boxes with a confidence above the threshold on each dataset. Comparing against real videos from the MOT challenge (red), it can be seen that UniCrowd performs similarly to the real videos, while LcrowdV (green) exhibits a rather different behavior.}
\label{fig:confidences}
\end{figure}

The graphical rendering is handled by Unity3D \cite{haas2014history}, using available assets for the generation of both the crowd and the environment. As shown in Fig. \ref{fig:visual_comparison}, available simulators like Agoraset \cite{allain2012agoraset} and LcrowdV \cite{cheung2016lcrowdv} also provide graphical rendering for their scenarios. However, Agoraset provides low quality rendering, especially as far as the environmental features is concerned. LCrowdV, instead, suffers of small image size and low illumination quality. A quantitative comparison in support to this, is shown in Fig. \ref{fig:confidences}, where it can be seen how the visual appearance of UniCrowd leads to a confidence distribution of human detection, which closely resembles the distribution on real videos with similar content \cite{milan2016mot16}. Other than improving the camera resolution and image size, our simulator provides multiple features to improve the visual fidelity, namely:

\begin{itemize}
    \item global dynamic illumination. It is possible to change the illumination depending on the time of the day (e.g. light fading and changing colors at sunset);
    \item weather conditions. Multiple different weather condition can be simulated, like rain and snow;
    \item camera modeling with real world distortion filters. Our rendering cameras are modeled to simulate the recording using real lenses (e.g. adding flares and barrel distortion) and real sensors (e.g. granular noise). Lenses, resolution, and focal length can be customised and can vary for each camera, simulating a real-world system where not all cameras are homogeneous and different cameras can be placed and used differently according to their characteristics;
    \item position and number of cameras;
    \item diverse agents appearance, in terms of gender, clothes, and body shape.
\end{itemize}

\subsection{Multi-label annotation}
\label{sec:multilabel}

\begin{table}[h]
\centering
\begin{tabular}{@{}lll@{}}
\toprule
\textbf{Low-level} & \textbf{Mid-level} & \textbf{Global} \\ \midrule
Bounding boxes               & Trajectories                  & Weather and time \\
Joints' position              & Anomalies                     & Crowd densities             \\
Segmentation masks         & People counting                & Spawn/target areas      \\ \bottomrule
\end{tabular}
\caption{Provided ground truth annotations for low and mid-level tasks, and global configuration parameters.}
\label{tab:annotations}
\end{table}

We provide ground truth data for multiple tasks, which refer to both appearance and behavioral features.

At the beginning of each simulation, we generate a ground truth file for each camera in the scene. At each time frame, for each agent visible by the camera, the ground truth file is filled with the annotations shown in Tab. \ref{tab:annotations}.

Among these data, some of them are strictly based on the behavior of pedestrians in the simulator (e.g trajectories and anomalies), while others rely purely on the visual appearance (e.g. segmentation masks and crowd counting). 

\section{Validation}
\label{sec:applications}

The provided ground truth data can be used for a variety of tasks/applications. For the validation of the proposed pipeline, we focus on people detection and segmentation (Fig. \ref{fig:yolo_on_sim}, Fig. \ref{fig:segmentation_data}), as they represent two key use-cases, on top of which, multiple scenarios can be envisaged.

The most common methods in literature target either robustness and accuracy of the detection and segmentation \cite{Detectron2018} or they explore the trade-off between accuracy and computational burden \cite{yolov3}.

\subsubsection{Dataset}

To test the performances of two off-the-shelf algorithms, and confirm their applicability on our synthetic data, we created a dataset, captured by multiple cameras in varying environmental conditions.

We used 3 different lighting setups corresponding to the natural illumination at 3 different times of the day (7:00, 12:00, 18:30), combined with 3 different levels of crowd density, each of them defined by the number of people generated at the spawn location at the beginning of each simulation (low = 40, medium=100, high=150).

Overall, we have a combination of 9 possible conditions, recorded by 5 different cameras\footnote{sample video sequences can be downloaded at
\href{https://drive.google.com/drive/folders/16Nb90YyOjfzDb9nolb6_efvl9syr6tAn?usp=sharing}{Crowd Simulations}}

Fig. \ref{fig:segmentation_data} shows a set of images and its associated segmentation masks taken randomly for each camera from the training dataset.
For all generated data, the ground truth is formatted as in COCO \cite{lin2014microsoft} using Run Length Encoding (RLE) to obtain the polygonal corresponding to the segmentation mask.

\subsubsection{Results}

In the detection domain \cite{metrics_detection}, we define for the evaluation a varying threshold of the Intersection over Union (IOU) at which we compute the $F1$ score. The $F1$ is used as the metric to evaluate the performances of the off-the-shelf versions of YOLO (detection) \cite{yolov3} and Detectron \cite{Detectron2018} on the synthetic data.

\begin{figure}[!ht]
\centering
\includegraphics[width=0.5\textwidth]{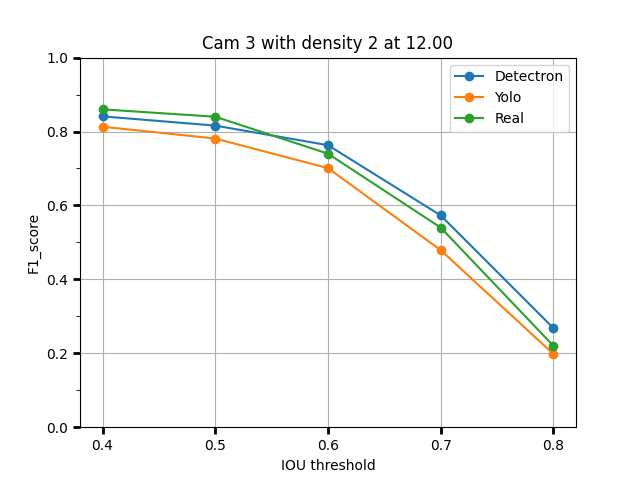}
\caption{F1-scores obtained varying the IOU threshold on the Cam3 at 12:00, with a medium density crowd with YOLO (orange line) and Detectron (blue line). YOLO \cite{amato2019learning} applied to real data \cite{dendorfer2019cvpr19} with similar viewpoint to the synthetic one (green line) performs similarly to simulated data (orange line). }
\label{fig:graficiIOU}
\end{figure}

In Fig. \ref{fig:graficiIOU}, we report a sample result for both YOLO and Detectron with a varying threshold between $[0.4,0.5,0.6,0.7,0.8]$.

\begin{figure}[!ht]
\centering
\subfigure[]{\includegraphics[width=.5\textwidth]{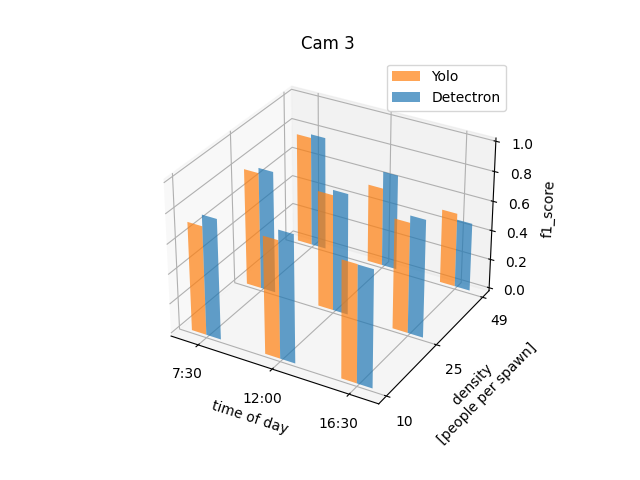}}
\subfigure[]{\includegraphics[width=.5\textwidth]{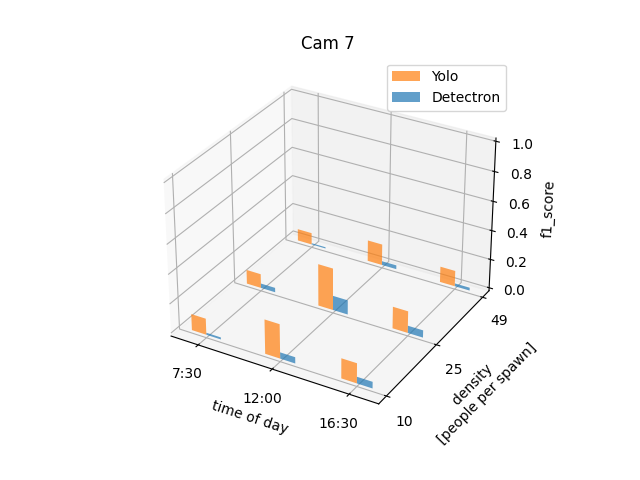}}
\caption{ F1-scores obtained applying off-the-shelf version of Detectron and YOLO on two different views ((a) Cam1, (b) Cam7) from synthetic data, varying the people density and the time of the day. Cam1 corresponds to a frontal viewpoint as shown as in Fig. \ref{fig:viewpoints}b; Cam7 corresponds to a quasi top-view as in Fig. \ref{fig:viewpoints}d, and shows how off-the-shelf algorithms struggle to obtain good results due to the lack of training data from such viewpoint.}
\label{fig:all_cams}
\end{figure}

In Fig. \ref{fig:all_cams}, we provide an overview of how the performances vary for two selected cameras, at different crowd densities and time of the day. Both algorithms obtain good results when dealing with the side-viewpoint (Fig. \ref{fig:all_cams}(a)), but struggle when dealing with the top-viewpoint (Fig. \ref{fig:all_cams}(b)). The top view-point presents the most challenging scenario for state-of-the-art algorithms, as shown in Fig. \ref{fig:viewpoints}, because of the scarcity of training data. Simulated data can then be of help with this respect, providing sufficient amount of data from any arbitrary viewpoint.
Overall, it can also be seen that illumination changes do not significantly influence the detection results; instead, performances tend to worsen when the density of the scene increases.

In the segmentation domain, we evaluate the performance of Detectron \cite{Detectron2018} using the standard COCO metrics \cite{lin2014microsoft}, which evaluates AP (Average Precision) for object detection and instance segmentation. For COCO, it is implied that AP is an average over all the categories, which is better referred as mean average precision (mAP).

The obtained results are reported in Tab. \ref{tab:segmentation_real_vs_synt}, showing comparable performances of the off-the-shelf model trained on real data and tested on both real and synthetic data.

\begin{figure}[!ht]
\centering
\subfigure[]{\includegraphics[width=.24\textwidth]{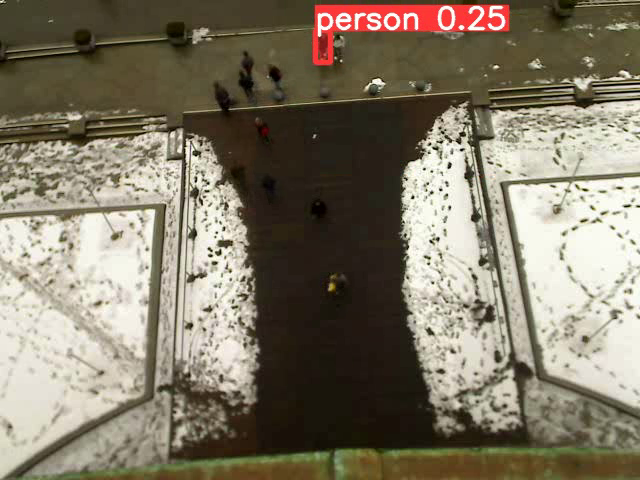}}
\subfigure[]{\includegraphics[width=.24\textwidth]{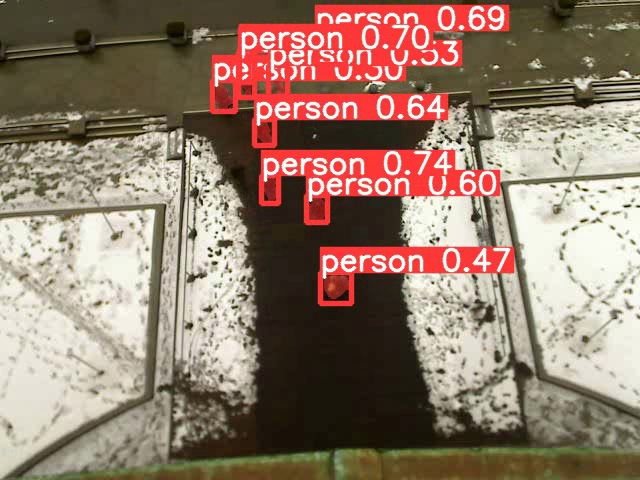}}
\subfigure[]{\includegraphics[width=.24\textwidth]{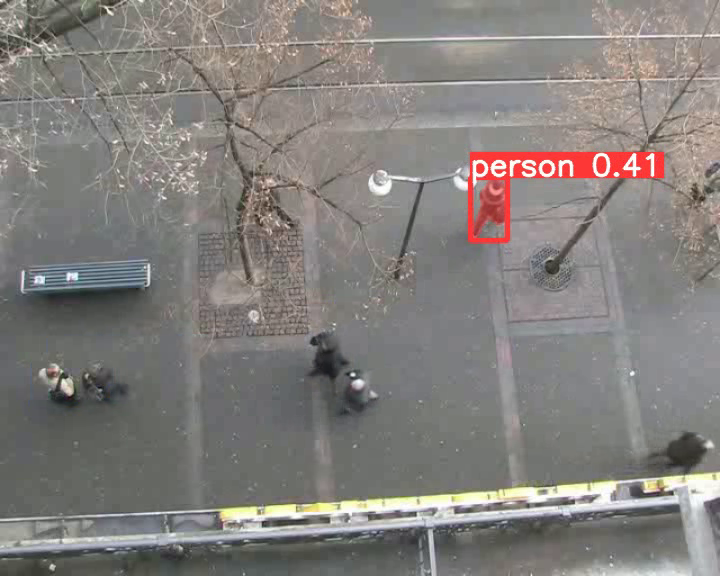}}
\subfigure[]{\includegraphics[width=.24\textwidth]{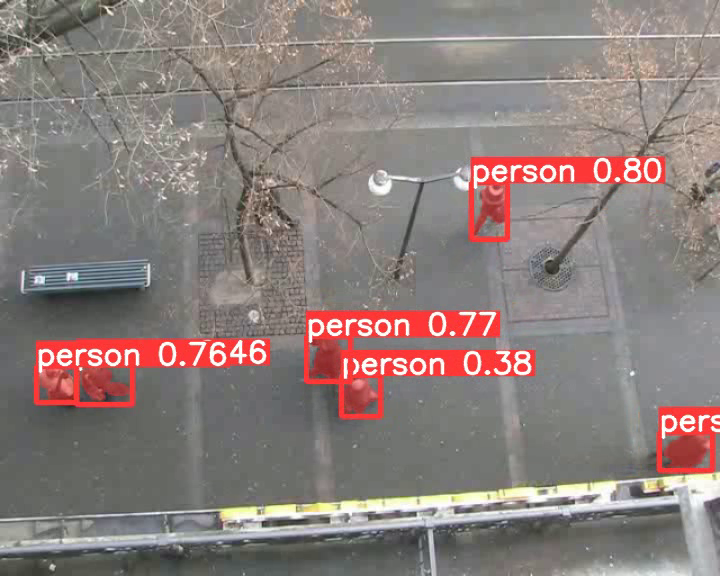}}
\caption{Challenging viewpoints detection can benefit from synthetic data augmentation. When applied on top-view images ((a) and (c)), the off-the-shelf version of YOLO struggles to identify people from the top view. Performing data augmentation with synthetic data provides much better results ((b) and (d)).}
\label{fig:yolo_augmented}
\end{figure}

Both detection and segmentation algorithms perform similarly on the synthetic video and in the real world, as shown in Fig. \ref{fig:segm_results}. In Fig. \ref{fig:yolo_augmented}, our experiments show how the missing viewpoint in the real datasets highly influences the results achievable in the synthetic domain (the algorithms cannot deal with top-view data), which can then be used to create new valuable training data.

\begin{table}[!ht]
    \centering
    \begin{tabular}{lcccc}
    \toprule
        & AP & APs & APm & APl \\
    \midrule
    Synthetic & 58.808 & 37.761 & 62.104 & 73.981 \\
    Real & 58.670 & 23.011 & 53.235 & 66.209 \\
    \bottomrule
    \end{tabular}
    \vspace{10px}
    \caption{Comparison of segmentation results obtained testing the off-the-shelf version of the Detectron algorithm on both real and synthetic test dataset.} (Keys: AP is the Average Precision, AP50 is the average Precision using a IoU threshold of 0.5, AP75 is the Average Precision using a IoU threshold of 0.75 , APs is the Average Precision on small objects , APm is the Average Precision on medium objects, APl is the Average Precision on large objects)
    \label{tab:segmentation_real_vs_synt}
\end{table}

\section{Applications}

The coming paragraphs are devoted to show how research in different areas of video surveillance and crowd monitoring can benefit from the proposed simulation platform. We present the use cases of trajectory prediction, anomaly detection, and human pose estimation.

\begin{figure}
    \centering
    \subfigure[ETH dataset]{\includegraphics[height=0.15\textwidth]{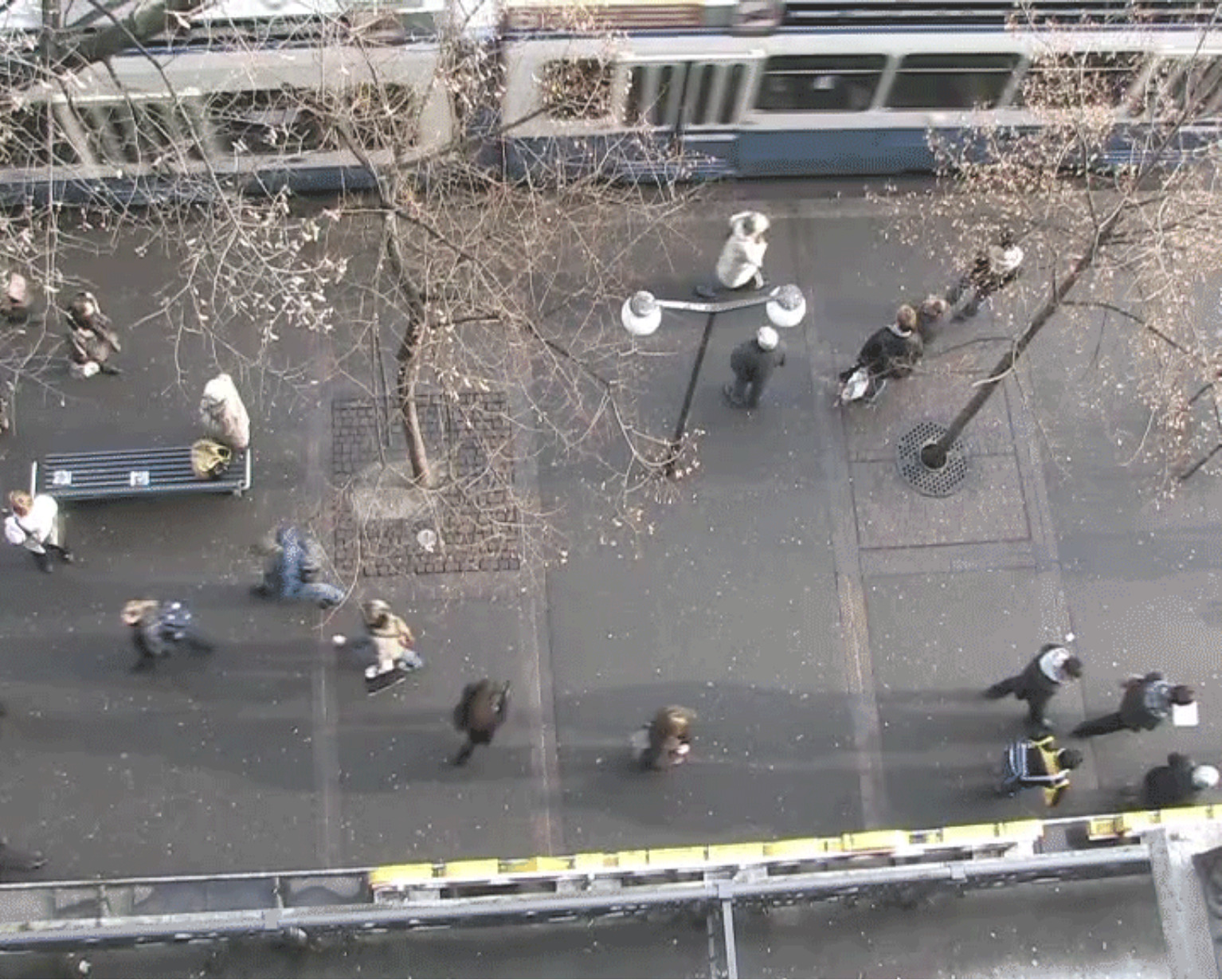}}
    \subfigure[Synthetic dataset]{\includegraphics[height=0.15\textwidth]{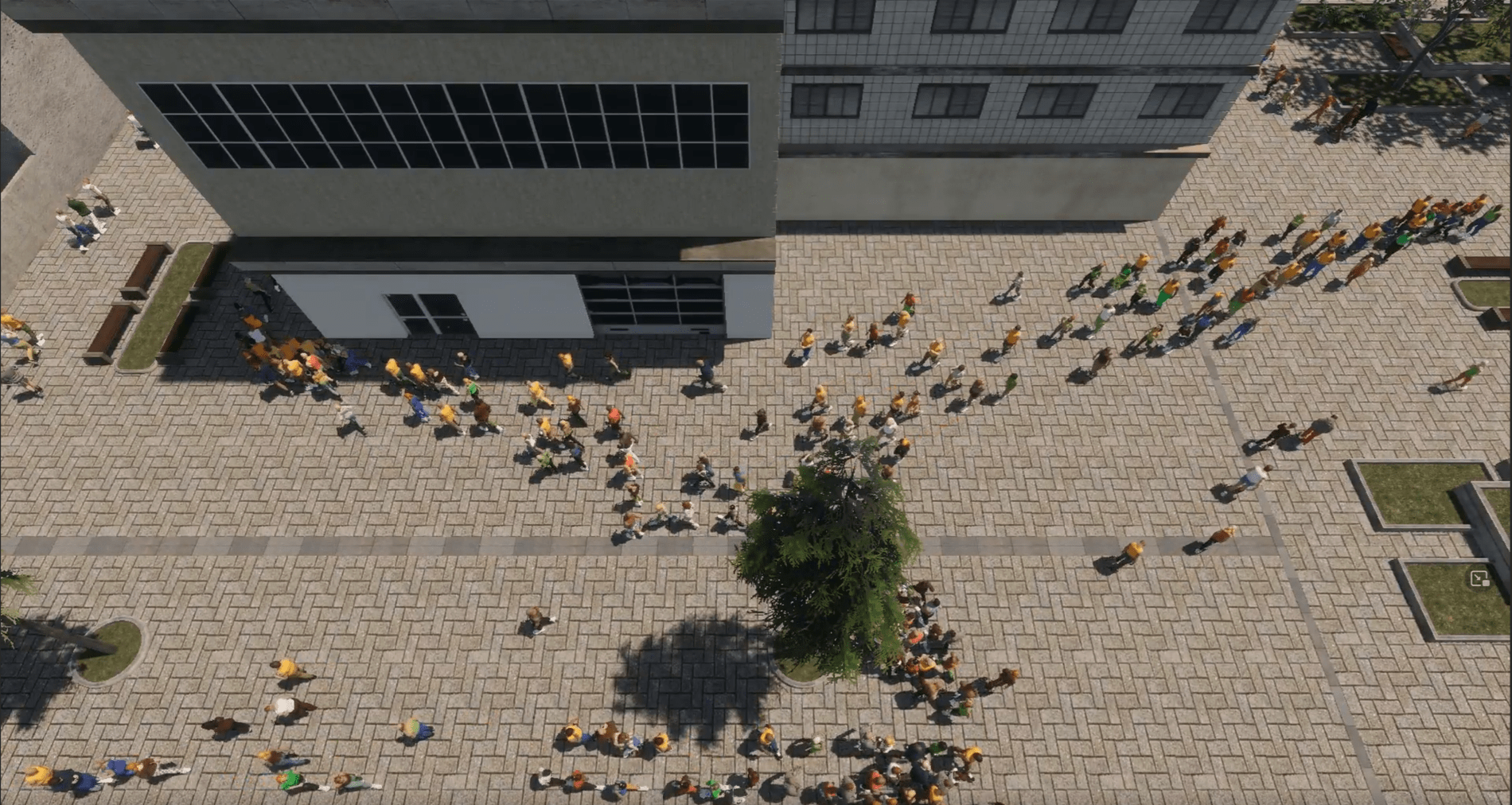}}
    \caption{A synthetic datasets allows for mimicking real scenes with longer trajectories, allowing for longer term predictions.}
    \label{fig:visual_comparison_traj}
\end{figure}
\textbf{Trajectory prediction}
With pedestrian detection and segmentation we focused on evaluating the \textit{visual fidelity} of the simulator; here, we aim instead on evaluating how each pedestrian moves in the environment, thus investigating the \textit{behavioral fidelity} of the simulation. A simulator allows generating longer trajectories compared to what is currently provided in the main reference datasets, as shown in Fig. \ref{fig:visual_comparison_traj}, making it possible to extend the temporal range of the prediction, usually confined within a little time interval of few seconds \cite{rudenko2020human}.

\textbf{Anomaly detection.}Anomalies are by definition rare events, and it therefore difficult to obtain a sufficient number of representative samples for detection and classification. 
Simulation can provide a mean to generate such anomalous situations, with the ability to scale up to a wide variety of samples. 

We tested an algorithm which has been successfully deployed on real videos \cite{ullah2014real} on the synthetic video generated by our simulator. From the experiments it turned out that the method, which is based on a Gaussian Mixture framework, performs similarly on synthetic and real videos, showing the effectiveness of the simulator to reproduce rare anomalies.

\textbf{Human pose estimation}
Human pose estimation (HPE) deals with capturing both the human position in the world and their pose at the joint-level. As an input, human pose estimation networks \cite{cao2019openpose} receive images or videos and as an output they produce 2D or 3D coordinates of ordered joints for each person in the frame. Similarly to the previous scenarios, manually annotating data for HPE is a demanding task, and requires a certain degree of expertise, other than the many hours required to obtain few seconds of reliable ground truth. Moreover, in some scenarios, real-world data (in particular 3D annotations) can only be considered to be pseudo-ground truth. For example, annotating human data from top-view images and videos cannot be considered accurate, because the position of the occluded parts of the body can barely be estimated.
Recent works \cite{garau2021deca} tackle this problem by relying on semi-synthetic data to improve HPE networks generalization from extreme viewpoints.
UniCrowd provides the possibility to automatically save multiple formats of human pose ground truth, both in 2D and in 3D, with many customization options (e.g., partial annotations in case of occlusions, different body models).

\section{Conclusions}
\label{sec:conclusions}

In this work, we have presented the multiple applications of simulation environments in addressing crowd analysis from a computer vision perspective. \textit{Visual fidelity} and \textit{behavioral fidelity} have been highlighted as a new paradigm that can be used to evaluate the effectiveness of synthetic data. Moreover, we have shown how a simulator that can generate data compliant with the real-world events can have a much broader impact than simple synthetic data based on scripted motion. We have chosen the use case of crowd analysis to display the potential of our framework, since it provides a challenging and complete example for both the behavioral and visual models. The results and the preliminary tests using state-of-the-art approaches have demonstrated the applicability of such a paradigm, and we firmly believe that the ever increasing capabilities of the simulation engines, will be key in the near future to guarantee the availability of sufficient data to tackle new research challenges.

{\small
\bibliographystyle{ieee_fullname}
\bibliography{egbib}

\begin{thebibliography}{10}\itemsep=-1pt

\bibitem{alahi2016social}
Alexandre Alahi, Kratarth Goel, Vignesh Ramanathan, Alexandre Robicquet, Li Fei-Fei, and Silvio Savarese.
\newblock Social lstm: Human trajectory prediction in crowded spaces.
\newblock In {\em Proceedings of the IEEE conference on computer vision and pattern recognition}, pages 961--971, 2016.

\bibitem{allain2012agoraset}
Pierre Allain, Nicolas Courty, and Thomas Corpetti.
\newblock Agoraset: a dataset for crowd video analysis.
\newblock 2012.

\bibitem{amato2019learning}
Giuseppe Amato, Luca Ciampi, Fabrizio Falchi, Claudio Gennaro, and Nicola Messina.
\newblock Learning pedestrian detection from virtual worlds.
\newblock In {\em International Conference on Image Analysis and Processing}, pages 302--312. Springer, 2019.

\bibitem{bahdanau2014neural}
Dzmitry Bahdanau, Kyunghyun Cho, and Yoshua Bengio.
\newblock Neural machine translation by jointly learning to align and translate.
\newblock {\em arXiv preprint arXiv:1409.0473}, 2014.

\bibitem{baslamisli2018joint}
Anil~S Baslamisli, Thomas~T Groenestege, Partha Das, Hoang-An Le, Sezer Karaoglu, and Theo Gevers.
\newblock Joint learning of intrinsic images and semantic segmentation.
\newblock In {\em Proceedings of the European Conference on Computer Vision (ECCV)}, pages 286--302, 2018.

\bibitem{bengio2012deep}
Yoshua Bengio.
\newblock Deep learning of representations for unsupervised and transfer learning.
\newblock In {\em Proceedings of ICML workshop on unsupervised and transfer learning}, pages 17--36. JMLR Workshop and Conference Proceedings, 2012.

\bibitem{Butler:ECCV:2012}
D.~J. Butler, J. Wulff, G.~B. Stanley, and M.~J. Black.
\newblock A naturalistic open source movie for optical flow evaluation.
\newblock In {A. Fitzgibbon et al. (Eds.)}, editor, {\em European Conf. on Computer Vision (ECCV)}, Part IV, LNCS 7577, pages 611--625. Springer-Verlag, Oct. 2012.

\bibitem{cao2019openpose}
Zhe Cao, Gines Hidalgo, Tomas Simon, Shih-En Wei, and Yaser Sheikh.
\newblock Openpose: realtime multi-person 2d pose estimation using part affinity fields.
\newblock {\em IEEE transactions on pattern analysis and machine intelligence}, 43(1):172--186, 2019.

\bibitem{chang2015shapenet}
Angel~X Chang, Thomas Funkhouser, Leonidas Guibas, Pat Hanrahan, Qixing Huang, Zimo Li, Silvio Savarese, Manolis Savva, Shuran Song, Hao Su, et~al.
\newblock Shapenet: An information-rich 3d model repository.
\newblock {\em arXiv preprint arXiv:1512.03012}, 2015.

\bibitem{chen2015deepdriving}
Chenyi Chen, Ari Seff, Alain Kornhauser, and Jianxiong Xiao.
\newblock Deepdriving: Learning affordance for direct perception in autonomous driving.
\newblock In {\em Proceedings of the IEEE international conference on computer vision}, pages 2722--2730, 2015.

\bibitem{cheung2016lcrowdv}
Ernest Cheung, Tsan~Kwong Wong, Aniket Bera, Xiaogang Wang, and Dinesh Manocha.
\newblock Lcrowdv: Generating labeled videos for simulation-based crowd behavior learning.
\newblock In {\em European Conference on Computer Vision}, pages 709--727. Springer, 2016.

\bibitem{curtis2016menge}
Sean Curtis, Andrew Best, and Dinesh Manocha.
\newblock Menge: A modular framework for simulating crowd movement.
\newblock {\em Collective Dynamics}, 1:1--40, 2016.

\bibitem{de2016detection}
Igor~R de Almeida, Vinicius~J Cassol, Norman~I Badler, Soraia~Raupp Musse, and Cl{\'a}udio~Rosito Jung.
\newblock Detection of global and local motion changes in human crowds.
\newblock {\em IEEE Transactions on Circuits and Systems for Video Technology}, 27(3):603--612, 2016.

\bibitem{de2017procedural}
C{\'e}sar~Roberto de Souza12, Adrien Gaidon, Yohann Cabon, and Antonio~Manuel L{\'o}pez.
\newblock Procedural generation of videos to train deep action recognition networks.
\newblock 2017.

\bibitem{fma_dataset}
Micha\"el Defferrard, Kirell Benzi, Pierre Vandergheynst, and Xavier Bresson.
\newblock {FMA}: A dataset for music analysis.
\newblock In {\em 18th International Society for Music Information Retrieval Conference (ISMIR)}, 2017.

\bibitem{dendorfer2019cvpr19}
Patrick Dendorfer, Hamid Rezatofighi, Anton Milan, Javen Shi, Daniel Cremers, Ian Reid, Stefan Roth, Konrad Schindler, and Laura Leal-Taixe.
\newblock Cvpr19 tracking and detection challenge: How crowded can it get?
\newblock {\em arXiv preprint arXiv:1906.04567}, 2019.

\bibitem{deng2009imagenet}
Jia Deng, Wei Dong, Richard Socher, Li-Jia Li, Kai Li, and Li Fei-Fei.
\newblock Imagenet: A large-scale hierarchical image database.
\newblock In {\em 2009 IEEE conference on computer vision and pattern recognition}, pages 248--255. Ieee, 2009.

\bibitem{dosovitskiy2017carla}
Alexey Dosovitskiy, German Ros, Felipe Codevilla, Antonio Lopez, and Vladlen Koltun.
\newblock Carla: An open urban driving simulator.
\newblock In {\em Conference on robot learning}, pages 1--16. PMLR, 2017.

\bibitem{erhan2010does}
Dumitru Erhan, Aaron Courville, Yoshua Bengio, and Pascal Vincent.
\newblock Why does unsupervised pre-training help deep learning?
\newblock In {\em Proceedings of the thirteenth international conference on artificial intelligence and statistics}, pages 201--208. JMLR Workshop and Conference Proceedings, 2010.

\bibitem{everingham2010pascal}
Mark Everingham, Luc Van~Gool, Christopher~KI Williams, John Winn, and Andrew Zisserman.
\newblock The pascal visual object classes (voc) challenge.
\newblock {\em International journal of computer vision}, 88(2):303--338, 2010.

\bibitem{fabbri21iccv}
Matteo Fabbri, Guillem Bras{\'o}, Gianluca Maugeri, Aljo{\v{s}}a O{\v{s}}ep, Riccardo Gasparini, Orcun Cetintas, Simone Calderara, Laura Leal-Taix{\'e}, and Rita Cucchiara.
\newblock Motsynth: How can synthetic data help pedestrian detection and tracking?
\newblock In {\em International Conference on Computer Vision (ICCV)}, 2021.

\bibitem{fabbri2018learning}
Matteo Fabbri, Fabio Lanzi, Simone Calderara, Andrea Palazzi, Roberto Vezzani, and Rita Cucchiara.
\newblock Learning to detect and track visible and occluded body joints in a virtual world.
\newblock In {\em Proceedings of the European conference on computer vision (ECCV)}, pages 430--446, 2018.

\bibitem{gaidon2018reasonable}
Adrien Gaidon, Antonio Lopez, and Florent Perronnin.
\newblock The reasonable effectiveness of synthetic visual data.
\newblock {\em International Journal of Computer Vision}, 126(9):899--901, 2018.

\bibitem{gaidon2016virtual}
Adrien Gaidon, Qiao Wang, Yohann Cabon, and Eleonora Vig.
\newblock Virtual worlds as proxy for multi-object tracking analysis.
\newblock In {\em Proceedings of the IEEE conference on computer vision and pattern recognition}, pages 4340--4349, 2016.

\bibitem{garau2021deca}
Nicola Garau, Niccol{\`o} Bisagno, Piotr Br{\'o}dka, and Nicola Conci.
\newblock Deca: Deep viewpoint-equivariant human pose estimation using capsule autoencoders.
\newblock {\em arXiv preprint arXiv:2108.08557}, 2021.

\bibitem{geiger2012we}
Andreas Geiger, Philip Lenz, and Raquel Urtasun.
\newblock Are we ready for autonomous driving? the kitti vision benchmark suite.
\newblock In {\em 2012 IEEE conference on computer vision and pattern recognition}, pages 3354--3361. IEEE, 2012.

\bibitem{Detectron2018}
Ross Girshick, Ilija Radosavovic, Georgia Gkioxari, Piotr Doll\'{a}r, and Kaiming He.
\newblock Detectron.
\newblock \url{https://github.com/facebookresearch/detectron}, 2018.

\bibitem{haas2014history}
John~K Haas.
\newblock A history of the unity game engine.
\newblock 2014.

\bibitem{hall1968proxemics}
Edward~T Hall, Ray~L Birdwhistell, Bernhard Bock, Paul Bohannan, A~Richard Diebold~Jr, Marshall Durbin, Munro~S Edmonson, JL Fischer, Dell Hymes, Solon~T Kimball, et~al.
\newblock Proxemics [and comments and replies].
\newblock {\em Current anthropology}, 9(2/3):83--108, 1968.

\bibitem{helbing1995social}
Dirk Helbing and Peter Molnar.
\newblock Social force model for pedestrian dynamics.
\newblock {\em Physical review E}, 51(5):4282, 1995.

\bibitem{henaff2021efficient}
Olivier~J H{\'e}naff, Skanda Koppula, Jean-Baptiste Alayrac, Aaron Van~den Oord, Oriol Vinyals, and Jo{\~a}o Carreira.
\newblock Efficient visual pretraining with contrastive detection.
\newblock In {\em Proceedings of the IEEE/CVF International Conference on Computer Vision}, pages 10086--10096, 2021.

\bibitem{ioffe2015batch}
Sergey Ioffe and Christian Szegedy.
\newblock Batch normalization: Accelerating deep network training by reducing internal covariate shift.
\newblock In {\em International conference on machine learning}, pages 448--456. PMLR, 2015.

\bibitem{ionescu2013human3}
Catalin Ionescu, Dragos Papava, Vlad Olaru, and Cristian Sminchisescu.
\newblock Human3. 6m: Large scale datasets and predictive methods for 3d human sensing in natural environments.
\newblock {\em IEEE transactions on pattern analysis and machine intelligence}, 36(7):1325--1339, 2013.

\bibitem{johnson2016driving}
Matthew Johnson-Roberson, Charles Barto, Rounak Mehta, Sharath~Nittur Sridhar, Karl Rosaen, and Ram Vasudevan.
\newblock Driving in the matrix: Can virtual worlds replace human-generated annotations for real world tasks?
\newblock {\em arXiv preprint arXiv:1610.01983}, 2016.

\bibitem{Joo_2017_TPAMI}
Hanbyul Joo, Tomas Simon, Xulong Li, Hao Liu, Lei Tan, Lin Gui, Sean Banerjee, Timothy~Scott Godisart, Bart Nabbe, Iain Matthews, Takeo Kanade, Shohei Nobuhara, and Yaser Sheikh.
\newblock Panoptic studio: A massively multiview system for social interaction capture.
\newblock {\em IEEE Transactions on Pattern Analysis and Machine Intelligence}, 2017.

\bibitem{kanazawa2018end}
Angjoo Kanazawa, Michael~J Black, David~W Jacobs, and Jitendra Malik.
\newblock End-to-end recovery of human shape and pose.
\newblock In {\em Proceedings of the IEEE conference on computer vision and pattern recognition}, pages 7122--7131, 2018.

\bibitem{kang2018beyond}
Di Kang, Zheng Ma, and Antoni~B Chan.
\newblock Beyond counting: comparisons of density maps for crowd analysis tasks—counting, detection, and tracking.
\newblock {\em IEEE Transactions on Circuits and Systems for Video Technology}, 29(5):1408--1422, 2018.

\bibitem{krahenbuhl2018free}
Philipp Kr{\"a}henb{\"u}hl.
\newblock Free supervision from video games.
\newblock In {\em Proceedings of the IEEE Conference on Computer Vision and Pattern Recognition}, pages 2955--2964, 2018.

\bibitem{krizhevsky2012imagenet}
Alex Krizhevsky, Ilya Sutskever, and Geoffrey~E Hinton.
\newblock Imagenet classification with deep convolutional neural networks.
\newblock {\em Advances in neural information processing systems}, 25:1097--1105, 2012.

\bibitem{lerner2007crowds}
Alon Lerner, Yiorgos Chrysanthou, and Dani Lischinski.
\newblock Crowds by example.
\newblock In {\em Computer Graphics Forum}, volume~26, pages 655--664. Wiley Online Library, 2007.

\bibitem{li2019recurrent}
Qiaozhe Li, Xin Zhao, Ran He, and Kaiqi Huang.
\newblock Recurrent prediction with spatio-temporal attention for crowd attribute recognition.
\newblock {\em IEEE Transactions on Circuits and Systems for Video Technology}, 30(7):2167--2177, 2019.

\bibitem{li2017paralleleye}
Xuan Li, Kunfeng Wang, Yonglin Tian, Lan Yan, and Fei-Yue Wang.
\newblock The paralleleye dataset: Constructing large-scale artificial scenes for traffic vision research.
\newblock {\em arXiv preprint arXiv:1712.08394}, 2017.

\bibitem{lin2014microsoft}
Tsung-Yi Lin, Michael Maire, Serge Belongie, James Hays, Pietro Perona, Deva Ramanan, Piotr Doll{\'a}r, and C~Lawrence Zitnick.
\newblock Microsoft coco: Common objects in context.
\newblock In {\em European conference on computer vision}, pages 740--755. Springer, 2014.

\bibitem{mayer2018makes}
Nikolaus Mayer, Eddy Ilg, Philipp Fischer, Caner Hazirbas, Daniel Cremers, Alexey Dosovitskiy, and Thomas Brox.
\newblock What makes good synthetic training data for learning disparity and optical flow estimation?
\newblock {\em International Journal of Computer Vision}, 126(9):942--960, 2018.

\bibitem{milan2016mot16}
Anton Milan, Laura Leal-Taix{\'e}, Ian Reid, Stefan Roth, and Konrad Schindler.
\newblock Mot16: A benchmark for multi-object tracking.
\newblock {\em arXiv preprint arXiv:1603.00831}, 2016.

\bibitem{moussaid2010walking}
Mehdi Moussa{\"\i}d, Niriaska Perozo, Simon Garnier, Dirk Helbing, and Guy Theraulaz.
\newblock The walking behaviour of pedestrian social groups and its impact on crowd dynamics.
\newblock {\em PloS one}, 5(4):e10047, 2010.

\bibitem{nikolenko2021synthetic}
Sergey~I Nikolenko et~al.
\newblock {\em Synthetic data for deep learning}.
\newblock Springer, 2021.

\bibitem{metrics_detection}
Rafael Padilla, Sergio~L. Netto, and Eduardo A.~B. da Silva.
\newblock A survey on performance metrics for object-detection algorithms.
\newblock In {\em 2020 International Conference on Systems, Signals and Image Processing (IWSSIP)}, pages 237--242, 2020.

\bibitem{pellegrini2009you}
Stefano Pellegrini, Andreas Ess, Konrad Schindler, and Luc Van~Gool.
\newblock You'll never walk alone: Modeling social behavior for multi-target tracking.
\newblock In {\em Computer Vision, 2009 IEEE 12th International Conference on}, pages 261--268. IEEE, 2009.

\bibitem{qiu2016unrealcv}
Weichao Qiu and Alan Yuille.
\newblock Unrealcv: Connecting computer vision to unreal engine.
\newblock In {\em European Conference on Computer Vision}, pages 909--916. Springer, 2016.

\bibitem{yolov3}
Joseph Redmon and Ali Farhadi.
\newblock Yolov3: An incremental improvement.
\newblock {\em arXiv}, 2018.

\bibitem{richter2017playing}
Stephan~R Richter, Zeeshan Hayder, and Vladlen Koltun.
\newblock Playing for benchmarks.
\newblock In {\em Proceedings of the IEEE International Conference on Computer Vision}, pages 2213--2222, 2017.

\bibitem{richter2016playing}
Stephan~R Richter, Vibhav Vineet, Stefan Roth, and Vladlen Koltun.
\newblock Playing for data: Ground truth from computer games.
\newblock In {\em European conference on computer vision}, pages 102--118. Springer, 2016.

\bibitem{robicquet2020learning}
A Robicquet, A Sadeghian, A Alahi, and S Savarese.
\newblock Learning social etiquette: Human trajectory prediction in crowded scenes.
\newblock In {\em European Conference on Computer Vision (ECCV)}, 2020.

\bibitem{ros2016synthia}
German Ros, Laura Sellart, Joanna Materzynska, David Vazquez, and Antonio~M Lopez.
\newblock The synthia dataset: A large collection of synthetic images for semantic segmentation of urban scenes.
\newblock In {\em Proceedings of the IEEE conference on computer vision and pattern recognition}, pages 3234--3243, 2016.

\bibitem{rudenko2020human}
Andrey Rudenko, Luigi Palmieri, Michael Herman, Kris~M Kitani, Dariu~M Gavrila, and Kai~O Arras.
\newblock Human motion trajectory prediction: A survey.
\newblock {\em The International Journal of Robotics Research}, 39(8):895--935, 2020.

\bibitem{sajid2020zoomcount}
Usman Sajid, Hasan Sajid, Hongcheng Wang, and Guanghui Wang.
\newblock Zoomcount: A zooming mechanism for crowd counting in static images.
\newblock {\em IEEE Transactions on Circuits and Systems for Video Technology}, 30(10):3499--3512, 2020.

\bibitem{savva2017minos}
Manolis Savva, Angel~X Chang, Alexey Dosovitskiy, Thomas Funkhouser, and Vladlen Koltun.
\newblock Minos: Multimodal indoor simulator for navigation in complex environments.
\newblock {\em arXiv preprint arXiv:1712.03931}, 2017.

\bibitem{shah2018airsim}
Shital Shah, Debadeepta Dey, Chris Lovett, and Ashish Kapoor.
\newblock Airsim: High-fidelity visual and physical simulation for autonomous vehicles.
\newblock In {\em Field and service robotics}, pages 621--635. Springer, 2018.

\bibitem{sheng2020hypothesis}
Hao Sheng, Yang Zhang, Yubin Wu, Shuai Wang, Weifeng Lyu, Wei Ke, and Zhang Xiong.
\newblock Hypothesis testing based tracking with spatio-temporal joint interaction modeling.
\newblock {\em IEEE Transactions on Circuits and Systems for Video Technology}, 30(9):2971--2983, 2020.

\bibitem{shorten2019survey}
Connor Shorten and Taghi~M Khoshgoftaar.
\newblock A survey on image data augmentation for deep learning.
\newblock {\em Journal of Big Data}, 6(1):1--48, 2019.

\bibitem{shotton2013real}
Jamie Shotton, Toby Sharp, Alex Kipman, Andrew Fitzgibbon, Mark Finocchio, Andrew Blake, Mat Cook, and Richard Moore.
\newblock Real-time human pose recognition in parts from single depth images.
\newblock {\em Communications of the ACM}, 56(1):116--124, 2013.

\bibitem{srivastava2014dropout}
Nitish Srivastava, Geoffrey Hinton, Alex Krizhevsky, Ilya Sutskever, and Ruslan Salakhutdinov.
\newblock Dropout: a simple way to prevent neural networks from overfitting.
\newblock {\em The journal of machine learning research}, 15(1):1929--1958, 2014.

\bibitem{sung2018learning}
Flood Sung, Yongxin Yang, Li Zhang, Tao Xiang, Philip~HS Torr, and Timothy~M Hospedales.
\newblock Learning to compare: Relation network for few-shot learning.
\newblock In {\em Proceedings of the IEEE conference on computer vision and pattern recognition}, pages 1199--1208, 2018.

\bibitem{treuille2006continuum}
Adrien Treuille, Seth Cooper, and Zoran Popovi{\'c}.
\newblock Continuum crowds.
\newblock {\em ACM Transactions on Graphics (TOG)}, 25(3):1160--1168, 2006.

\bibitem{ullah2014real}
Habib Ullah, Mohib Ullah, and Nicola Conci.
\newblock Real-time anomaly detection in dense crowded scenes.
\newblock In {\em Video Surveillance and Transportation Imaging Applications 2014}, volume 9026, pages 51--57. SPIE, 2014.

\bibitem{van2008reciprocal}
Jur Van~den Berg, Ming Lin, and Dinesh Manocha.
\newblock Reciprocal velocity obstacles for real-time multi-agent navigation.
\newblock In {\em 2008 IEEE international conference on robotics and automation}, pages 1928--1935. Ieee, 2008.

\bibitem{wu2016google}
Yonghui Wu, Mike Schuster, Zhifeng Chen, Quoc~V Le, Mohammad Norouzi, Wolfgang Macherey, Maxim Krikun, Yuan Cao, Qin Gao, Klaus Macherey, et~al.
\newblock Google's neural machine translation system: Bridging the gap between human and machine translation.
\newblock {\em arXiv preprint arXiv:1609.08144}, 2016.

\bibitem{xian2017zero}
Yongqin Xian, Bernt Schiele, and Zeynep Akata.
\newblock Zero-shot learning-the good, the bad and the ugly.
\newblock In {\em Proceedings of the IEEE Conference on Computer Vision and Pattern Recognition}, pages 4582--4591, 2017.

\bibitem{xie2016semantic}
Jun Xie, Martin Kiefel, Ming-Ting Sun, and Andreas Geiger.
\newblock Semantic instance annotation of street scenes by 3d to 2d label transfer.
\newblock In {\em Proceedings of the IEEE Conference on Computer Vision and Pattern Recognition}, pages 3688--3697, 2016.

\bibitem{Yu2005}
W.~J. Yu, R. Chen, L.~Y. Dong, and S.~Q. Dai.
\newblock Centrifugal force model for pedestrian dynamics.
\newblock {\em Phys. Rev. E}, 72:026112, Aug 2005.

\bibitem{yuan2020deep}
Qiangqiang Yuan, Huanfeng Shen, Tongwen Li, Zhiwei Li, Shuwen Li, Yun Jiang, Hongzhang Xu, Weiwei Tan, Qianqian Yang, Jiwen Wang, et~al.
\newblock Deep learning in environmental remote sensing: Achievements and challenges.
\newblock {\em Remote Sensing of Environment}, 241:111716, 2020.

\bibitem{zhang2021weakly}
Dingwen Zhang, Junwei Han, Gong Cheng, and Ming-Hsuan Yang.
\newblock Weakly supervised object localization and detection: A survey.
\newblock {\em IEEE transactions on pattern analysis and machine intelligence}, 2021.

\bibitem{zhang2019comprehensive}
Hong-Bo Zhang, Yi-Xiang Zhang, Bineng Zhong, Qing Lei, Lijie Yang, Ji-Xiang Du, and Duan-Sheng Chen.
\newblock A comprehensive survey of vision-based human action recognition methods.
\newblock {\em Sensors}, 19(5):1005, 2019.

\bibitem{zhou2021review}
S~Kevin Zhou, Hayit Greenspan, Christos Davatzikos, James~S Duncan, Bram Van~Ginneken, Anant Madabhushi, Jerry~L Prince, Daniel Rueckert, and Ronald~M Summers.
\newblock A review of deep learning in medical imaging: Imaging traits, technology trends, case studies with progress highlights, and future promises.
\newblock {\em Proceedings of the IEEE}, 2021.

\end{thebibliography}
}

\end{document}